\documentclass{tlp}
\usepackage[all]{xy}
\usepackage{url}
\usepackage{graphicx}
\usepackage{latexsym}
\usepackage{algorithm}
\usepackage{algpseudocode}
\usepackage{subfigure}
\usepackage{multirow}
\usepackage{pbox}

\newcommand{\emalgorithm}{EMBLEM}
\newcommand{\emsalgorithm}{SLIPCOVER}
\newtheorem{definition}{Definition} %
\newtheorem{example}{Example} %

\newcommand\Refs{{\mathit{Refs}}}

\submitted{09 April 2012}
\revised{14 October 2012 / 18 May 2013}
\accepted{07 August 2013}

\begin{document}

\title[Learning Probabilistic Logic Programs]{Structure Learning of Probabilistic Logic Programs by Searching the Clause Space}
\author[E. Bellodi and F. Riguzzi] {ELENA BELLODI $^1$ and FABRIZIO RIGUZZI $^2$ \\
$^1$ Dipartimento di Ingegneria -- University of Ferrara\\
Via Saragat 1, 44122, Ferrara, Italy  \\
$^2$ Dipartimento di Matematica e Informatica -- University of Ferrara\\
Via Saragat 1, 44122, Ferrara, Italy  \\
\email{elena.bellodi@unife.it,fabrizio.riguzzi@unife.it}}

\pagerange{\pageref{firstpage}--\pageref{lastpage}}
\setcounter{page}{1}
\maketitle

\label{firstpage}

\noindent
{\bf Note}: This article will appear in Theory and Practice of Logic Programming, 
\copyright Cambridge University Press.

\begin{abstract}
Learning probabilistic logic programming languages is receiving an increasing attention and  systems are available for learning the parameters (PRISM, LeProbLog, LFI-ProbLog
and EMBLEM) or both the structure and the parameters (SEM-CP-logic and SLIPCASE) of these languages. 
In this paper we present the algorithm SLIPCOVER for ``Structure LearnIng of Probabilistic logic programs by searChing OVER the clause space''. It performs a beam search in the space of probabilistic clauses and a greedy search in the space of theories, using the log likelihood of the data as the guiding heuristics. To estimate the log likelihood SLIPCOVER performs Expectation Maximization with EMBLEM.
The algorithm has been tested on five real world datasets and compared with SLIPCASE, SEM-CP-logic, Aleph and two algorithms for learning Markov Logic Networks (Learning using Structural Motifs (LSM) and ALEPH++ExactL1).  SLIPCOVER achieves higher areas under the precision-recall and ROC curves in most cases.
\end{abstract} 

\begin{keywords}
Probabilistic Inductive Logic Programming, Statistical Relational Learning, Structure Learning, Distribution Semantics, Logic Programs with Annotated Disjunction, CP-Logic 
\end{keywords}

\section{Introduction}
\label{intro}
Recently much work in Machine Learning has concentrated on representation languages able to combine aspects of logic and probability, leading to the birth of a whole field called Statistical Relational Learning (SRL). The ability to model both \textit{complex} and \textit{uncertain} relationships among entities is very important for learning accurate models of many domains. The standard frameworks for handling these features are first-order logic and probability theory respectively. 
Thus we would like to be able to learn and perform inference in languages that integrate the two, unlike traditional Inductive Logic Programming (ILP) methods which only address the complexity issue.

Probabilistic Logic Programming (PLP) has recently received an increasing attention for its ability to incorporate probability in logic programming. 
Among various proposals for PLP, the one based on the distribution semantics \cite{DBLP:conf/iclp/Sato95} is
gaining popularity and is the basis for languages such as Probabilistic Logic Programs \cite{DBLP:conf/lpar/Dantsin91}, Probabilistic Horn Abduction  \cite{DBLP:journals/ngc/Poole93}, PRISM~\cite{DBLP:conf/iclp/Sato95},
 Independent Choice Logic  \cite{Poo97-ArtInt-IJ}, pD \cite{DBLP:journals/jasis/Fuhr00}, Logic Programs with Annotated Disjunctions (LPADs) \cite{VenVer04-ICLP04-IC}, ProbLog \cite{DBLP:conf/ijcai/RaedtKT07} and CP-logic \cite{DBLP:journals/tplp/VennekensDB09}.

Inference for PLP languages can be performed with a number of algorithms, which in many cases find explanations for queries and compute their probability by building a Binary Decision Diagram (BDD) \cite{DBLP:conf/ijcai/RaedtKT07,Rig-AIIA07-IC,DBLP:journals/tplp/KimmigDRCR11,RigSwi11-ICLP11-IJ}.

Various works have started to appear on the problem of learning the parameters of PLP languages under the distribution semantics: 
LeProbLog \cite{DBLP:conf/pkdd/GutmannKKR08} uses gradient descent while LFI-ProbLog
  \cite{DBLP:conf/pkdd/GutmannTR11} and \emalgorithm{} \cite{BelRig12-IA-IJ,BelRig13-IDA-IJ} use an Expectation Maximization approach where  the expectations are computed directly from BDDs. 

The problem of learning the structure of these languages is also becoming of interest, with works such as  \cite{DBLP:journals/ml/RaedtKKRT08}, where  a theory compression algorithm for ProbLog is presented, and 
 \cite{DBLP:journals/fuin/MeertSB08}, where ground LPADs are learned using Bayesian Networks techniques. 
SLIPCASE \cite{BelRig12-ILP11-IC} also learns the structure of LPADs by
performing a beam search in the space of probabilistic theories using the log likelihood (LL) of the data as the guiding heuristics. To estimate the LL,  it performs a limited number of Expectation Maximization iterations of \emalgorithm{}. 

The structure learning task may be addressed with a discriminative or generative approach. A discriminative learning problem is characterized by specific target predicate(s) that must be predicted. The search for  clauses is directly guided by the goal of maximizing the predictive accuracy of the resulting theory on the target predicates. 
A generative learner attempts, on the contrary, to learn a theory that is equally capable of predicting the truth value of all predicates.

In this paper we propose an evolution of SLIPCASE called SLIPCOVER, 
for ``Structure LearnIng of Probabilistic logic programs by searChing OVER the clause space''. SLIPCASE is based on a  simple search strategy that iteratively performs theory revision. Differently from it, SLIPCOVER first searches the space of clauses storing all the promising ones, dividing them into clauses for target predicates (those we want to predict) and clauses for background predicates (the remaining ones), with a discriminative approach. This search starts from a set of ``bottom clauses'' generated as in Progol \cite{DBLP:journals/ngc/Muggleton95} and looks for good refinements in terms of LL.
Then it performs a greedy search in the space of theories, by trying to add each clause for a target predicate to the current theory. Finally, it performs parameter learning with EMBLEM on the best target theory plus the clauses for background predicates. \emsalgorithm{}  can learn general LPADs including non\--ground programs.  

Finally, Markov Logic Networks (MLNs) are a recently developed SRL model that generalizes both first order logic and Markov networks \cite{DBLP:journals/ml/RichardsonD06}, for which several parameter and structure learning algorithms have been proposed.

The aim of the paper is to demonstrate that a system based on ILP and PLP is competitive or superior to existing purely ILP or SRL methods. Moreover, the paper shows how the improved search strategy implemented in SLIPCOVER produces superior results with respect to the simpler SLIPCASE. 

The paper is organized as follows. Section \ref{LPAD} presents Probabilistic Logic Programming, concentrating on LPADs.  Section \ref{emblem} describes \emalgorithm{}  in details.  Section \ref{slipcover} illustrates \emsalgorithm{} while Section \ref{related} discusses related work. In Section \ref{experiments} we present the results of the experiments. Section \ref{conc} concludes the paper and proposes directions for future work.

\section{Probabilistic Logic Programming}
\label{LPAD}
The distribution semantics \cite{DBLP:conf/iclp/Sato95} is one of the most interesting approaches to the integration of logic programming and probability. It was introduced for the PRISM language but is shared by 
many other languages. 
 A program in one of these languages defines a probability distribution over normal logic programs called \emph{instances}. Each normal program is assumed to have a total well-founded model \cite{well-founded} thus each program can be associated with a Herbrand interpretation (a \emph{world}) that is its model and the distribution over instances directly translates into a distribution over Herbrand interpretations. Then, the distribution is extended to queries and the probability of a query is obtained by marginalizing the joint distribution of the query and the programs.
The languages following the distribution semantics differ in the way they define the distribution over logic programs but  have the same expressive power: there are transformations with linear complexity that can convert each one into the others \cite{VenVer03-TR,DeR-NIPS08}. 
In this paper we will use LPADs for their general syntax. We review here the semantics in the case of no function symbols for the sake of simplicity.

In LPADs the alternatives are encoded in the head of clauses in the form of a disjunction in which each atom is annotated with a probability. 
Formally a \emph{Logic Program with Annotated Disjunctions} $T$ consists of a finite set of
annotated disjunctive clauses \cite{VenVer04-ICLP04-IC}. An annotated disjunctive clause $C_i$ is 
of the form
$$h_{i1}:\Pi_{i1}; \ldots ; h_{in_i}:\Pi_{in_i} \ {:\!-}\   b_{i1},\ldots,b_{im_i}$$
In
such a clause, $h_{i1},\ldots, h_{in_i}$ are  logical atoms,  $b_{i1},\ldots ,b_{im_i}$ are logical literals and   $\{\Pi_{i1},\ldots,\Pi_{in_i}\}$ are real numbers in the interval $[0,1]$
such that $\sum_{k=1}^{n_i} \Pi_{ik}\leq 1$; $b_{i1},\ldots,b_{im_i}$ is indicated with $body(C_i)$.  Note that, if $n_i=1$ and $\Pi_{i1} = 1$, the clause corresponds to a non-disjunctive clause.
If $\sum_{k=1}^{n_i} \Pi_{ik}<1$, the head of the annotated disjunctive clause implicitly contains an extra atom $null$ that does not appear in the body of any clause and whose annotation is $1-\sum_{k=1}^{n_i} \Pi_{ik}$. We denote by $ground(T)$ the grounding of an LPAD $T$.

An  \emph{atomic choice} \cite{Poo97-ArtInt-IJ} is a triple $(C_i,\theta_j,k)$ where $C_i\in T$, $\theta_j$ is a substitution that grounds $C_i$ and $k\in\{1,\ldots,n_i\}$ identifies one of the head atoms. 
In practice $C_i\theta_j$ corresponds to a random variable $X_{ij}$ and an atomic choice $(C_i,\theta_j,k)$ to an assignment $X_{ij}=k$.
 A set of atomic choices $\kappa$ is \emph{consistent} if 
 only one head is selected from the same ground clause; we assume independence between the different choices. A  \emph{composite choice} $\kappa$ is a consistent set of atomic choices \cite{Poo97-ArtInt-IJ}. 
The \emph{probability $P(\kappa)$ of a composite choice} $\kappa$ is the product of the probabilities of
the independent atomic choices, i.e.
$P(\kappa)=\prod_{(C_i,\theta_j,k)\in \kappa}\Pi_{ik}$.

A \emph{selection} $\sigma$ is a  composite choice that, for each  clause $C_i\theta_j$ in $ground(T)$, contains an atomic choice 
$(C_i,\theta_j,k)$. Let us indicate with $S_T$ the set of all selections.
A selection $\sigma$ identifies a  normal logic program  $l_\sigma$ defined as $l_\sigma=\{(h_{ik}\leftarrow  body(C_i))\theta_j| (C_i,\theta_j,k)\in \sigma\}$.
$l_\sigma$ is called an \textit{instance} of $T$.
Since selections are composite choices, we can assign a probability to  instances: $P(l_\sigma)=P(\sigma)=\prod_{(C_i,\theta_j,k)\in \sigma}\Pi_{ik}$. 

 We consider only {\em sound} LPADs as defined below.
\begin{definition} 
An LPAD $T$ is called $sound$ iff for each selection $\sigma$ in $S_T$, the well-founded model of the program $l_\sigma$ chosen by $\sigma$ is two-valued.
\end{definition}
A particularly loose sufficient  condition for the soundness of LPADs is the bounded term-size property, defined in \cite{RigSwi11-TPLP-IJ},
which is based on a characterization of the well-founded semantics in terms of
an iterated fixpoint \cite{DBLP:conf/pods/Przymusinski89}. A bounded term-size program is such that
in each iteration of the fixpoint the size of true atoms does not grow indefinitely.
LPAD without negation in clauses' bodies are sound, as the well-founded model coincides with the least Herbrand model.

We write $l_\sigma\models Q$ to mean that the query $Q$ is true in the well\--founded model of the program $l_\sigma$.

We denote the set of all instances  by $L_T$.
A composite choice $\kappa$ identifies a set of instances $\lambda_\kappa=\{l_\sigma|\sigma \in S_T, \sigma \supseteq \kappa\}$.
We define the set of instances identified by a set of composite choices $K$ as
$\lambda_K=\bigcup_{\kappa\in K}\lambda_\kappa$.

Let $P(L_T)$ be the distribution over instances. The probability of a query $Q$ given an instance $l$ is
 $P(Q|l)=1$ if $l\models Q$ and 0 otherwise.
The probability of a query $Q$ is given by
\begin{equation}
\label{ds}
P(Q)=\sum_{l\in L_T}P(Q,l)=\sum_{l \in L_T}P(Q|l)P(l)=\sum_{l\in L_T: l\models Q}P(l)
\end{equation}

\begin{example} \label{stromboli}
 The following LPAD $T$ is inspired by the morphological characteristics of the Stromboli Italian island:

\vspace{0.2cm}
$
 \begin{array}{llll}
C_1&=&eruption:0.6\ ;\ earthquake:0.3 \ {:\!-} \  sudden\_energy\_release,\\
&& fault\_rupture(X).\\ 
C_2&=&sudden\_energy\_release:0.7.\\
C_3&=&fault\_rupture(southwest\_northeast).\\
C_4&=&fault\_rupture(east\_west).
 \end{array}$

\vspace{0.2cm}
\noindent
The Stromboli island is located at the intersection of two geological faults, one in the southwest-northeast direction, the other in the east-west direction, and contains one of the three volcanoes that are active  in Italy.
This program models the possibility that an eruption or an earthquake  occurs at Stromboli. If there is a sudden energy release under the island and there is a fault rupture ($C_1$), then there can be an eruption of the volcano on the island with probability 0.6 or an earthquake in the area with probability 0.3 or no event
 (the implicit $null$ atom) with probability 0.1. %
The energy release occurs with probability 0.7 while we are sure that ruptures occur in both faults.

Clause $C_1$ has two groundings, $C_1\theta_1$ with $\theta_1=\{X/southwest\_northeast\}$ and $C_1\theta_2$ with $\theta_2=\{X/east\_west\}$, so there are two  random variables $X_{11}$ and $X_{12}$. Clause $C_2$  has only one grounding $C_2\emptyset$ instead, so there is one  random variable $X_{21}$. $X_{11}$ and $X_{12}$ can take on three values since $C_1$ has three head atoms; similarly $X_{21}$ can take on two values since $C_2$ has two head atoms. 
$T$ has 18 instances, the query $eruption$ is true in 5 of them and its probability is 

\noindent 
$P(eruption)=0.6\cdot 0.6\cdot 0.7+0.6\cdot 0.3\cdot 0.7+0.6\cdot 0.1\cdot 0.7+0.3\cdot 0.6\cdot 0.7+0.1\cdot 0.6\cdot 0.7=0.588.$
For instance, the second term of $P(eruption)$ corresponds to the following instance of  $T$:

\vspace{0.2cm}
$
 \begin{array}{llll}
C_{11}&=&eruption{:\!-} \  sudden\_energy\_release,\\
&& fault\_rupture(southwest\_northeast).\\ 
C_{12}&=&earthquake {:\!-}  sudden\_energy\_release,fault\_rupture(east\_west).\\
C_2&=&sudden\_energy\_release.\\
C_3&=&fault\_rupture(southwest\_northeast).\\
C_4&=&fault\_rupture(east\_west).
 \end{array}$
\end{example}
As this example shows, multiple-head atoms  are particularly useful when clauses have a causal interpretation: the body represents an event that, when happening, has a random consequence among those in the head.
In addition, note that this LPAD  allows the  events \textit{eruption} and \textit{earthquake} to occurr at the same time, by virtue of the multiple groundings of clause $C_1$, as the above instance shows.

The semantics  associates one random variable with every grounding of a clause. In some domains, this may result in too many random variables. In order to contain the number of variables and thus simplify inference, we may introduce an approximation at the level of the instantiations, by grounding \textit{only some} of the variables of the clauses, at the expenses of the accuracy in modeling the domain.
A typical compromise between accuracy and complexity is to consider the grounding  of variables \textit{in the head only}: in this way, a ground atom entailed by two separate ground instances of a clause is assigned the same probability, all other things being equal, of a ground atom entailed by a single ground clause, while in the ``standard'' semantics the first would have a larger probability, as more evidence is available for its entailment.
This ``approximate'' semantics can be interpreted as stating that a ground atom is entailed by a clause with the probability given by its annotation if there is at least one substitution for the variables appearing only in the body  such that the body is true.
We have adopted this semantics in some experiments with SLIPCOVER and SLIPCASE in Section \ref{experiments}.

\begin{example}[Example \ref{stromboli} cont.] \label{stromboli-simp}
In the approximate semantics, $C_1$  is associated with a single random variable $X_{11}$.  In this case $T$ has 6 instances, the query $eruption$ is true in 1 of them and its probability is $P(eruption)=0.6\cdot0.7=0.42.$
So $eruption$ is assigned a lower probability with respect to the standard semantics because the two independent groundings of clause $C_1$, differing in the fault name, are not considered separately.
\end{example}
In practice inference algorithms find \emph{explanations} for a query: a composite choice $\kappa$ is an \emph{explanation} for a query $Q$ if $Q$ is entailed  by every instance of $\lambda_\kappa$. 
In particular, algorithms find a covering set of explanations for the query, where
a set of composite choices $K$ is \emph{covering} with respect to $Q$ if every program $l_\sigma$ in which $Q$ is entailed is in $\lambda_K$.
The problem of computing the probability of a query $Q$ can thus be reduced to computing the probability of the 
Boolean function 
\begin{equation}\label{dnf}
f_Q(\mathbf{X})=\bigvee_{\kappa\in E(Q)}\bigwedge_{(C_i,\theta_j,k)\in \kappa}X_{ij}=k
\end{equation}
where $E(Q)$  is a covering set of explanations for $Q$. 
\begin{example}[Example \ref{stromboli} cont.] \label{stromboli-exp}
The query $eruption$ has the covering set of explanations $E(eruption)=\{\kappa_1, \kappa_2\}$ where:
$$
\begin{array}{llll} \label{explanations}
\kappa_1&=&\{(C_1,\{X/southwest\_northeast\},1),(C_2,\{\},1)\}\\ \kappa_2&=&\{(C_1,\{X/east\_west\},1),(C_2,\{\},1)\} 
\end{array}
$$
Each atomic choice $(C_i, \theta_j, k)$ is represented by the propositional equation $X_{ij} = k$:
$$
\begin{array}{llll}
(C_1,\{X/southwest\_northeast\},1) &\rightarrow& X_{11} = 1 \\
(C_1,\{X/east\_west\},1) &\rightarrow& X_{12} = 1\\
(C_2,\{\},1) &\rightarrow& X_{21} = 1
 \end{array}
$$
The resulting Boolean function $f_{eruption}(\mathbf X)$ takes on value 1 if the values of the variables correspond to an explanation for the goal.
Equations for a single explanation are conjoined and the conjunctions for the different explanations are disjoined.
The set of explanations $E(eruption)$ can thus be encoded by the function:
\begin{equation}
\label{Feruption}
f_{eruption}(\mathbf{X})=(X_{11} = 1 \wedge  X_{21} = 1) \vee (X_{12} = 1 \wedge X_{21} = 1)
\end{equation}
\end{example}
Explanations however, differently from instances, are not necessarily mutually exclusive with respect to each other, so the probability of the query can not be computed by a summation as in (\ref{ds}).  %
In fact, computing the probability of a formula in Disjunctive Normal Form was shown to be \#P-hard  \cite{RCDB03}.

Various techniques have then been proposed for solving the inference problem in an exact or approximate way: using Multivalued Decision Diagrams (MDDs) \cite{DBLP:conf/ijcai/RaedtKT07,Rig-AIIA07-IC,Rig09-LJIGPL-IJ,RigSwi10-ICLP10-IC,DBLP:journals/tplp/KimmigDRCR11,RigSwi11-ICLP11-IJ,RigSwi11-TPLP-IJ}, modifying SLG resolution 
\cite{Rig08-ICLP08-IC,Rig10-FI-IJ}, exploiting specific conditions \cite{DBLP:journals/jair/SatoK01,Rig13-CJ-IJ} or using a Monte Carlo approach \cite{DBLP:journals/tplp/KimmigDRCR11,BraRig10-ILP10-IC,Rig13-FI-IJ}.

Here we consider the approach based on MDDs since it was shown to perform exact inference for general probabilistic logic programs effectively.

An MDD \cite{ThaDav-MDD-78} represents a  function $f(\mathbf{X})$ taking Boolean values on a set of multivalued variables ${\mathbf X}$ by means of  
a rooted graph that has one level for each variable. Each node is associated with the variable of its level and has one child for each possible value of the variable. The leaves store either 0 or 1.
Given values for all the variables ${\mathbf X}$, we can compute the value of $f(\mathbf{X})$  by traversing the graph starting from the root and returning the value associated with the leaf that is reached.
MDDs can be built by combining simpler MDDs using Boolean operators. While building MDDs, simplification operations can be applied that merge or delete nodes. Merging is performed when the diagram contains two identical sub-diagrams, while deletion is performed when all arcs from a node  point to the same node.
In this way a reduced MDD is obtained, often with a much smaller number of nodes with respect to the original MDD.

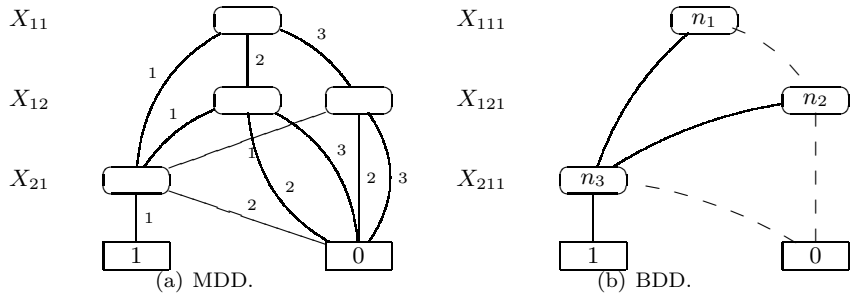
\begin{figure}[ht]
\centering
\subfigure
	[MDD.\label{MDD}]	{

$$
\xymatrix@=6mm
{ X_{11}& & *=<25pt,10pt>[F-:<3pt>]{\ }
\ar@/_1pc/@{-}[ldd]_1 \ar@{-}[d]^2 \ar@/^/@{-}[dr]^3\\ 
X_{12} & & *=<25pt,10pt>[F-:<3pt>]{\ } 
\ar@/_/@{-}[dl]_1\ar@/_1pc/@{-}[ddr]^2 \ar@/^1pc/@{-}[ddr]^3
& *=<25pt,10pt>[F-:<3pt>]{\ } 
\ar@{-}[dll]^1\ar@{-}[dd]^2 \ar@/^1pc/@{-}[dd]^3\\
X_{21}& *=<25pt,10pt>[F-:<3pt>]{\ }
\ar@{-}[d]^1\ar@{-}[drr]^2  \\
&*=<25pt,10pt>[F]{1} &&*=<25pt,10pt>[F]{0}
}
$$
}
\subfigure
	[BDD.\label{BDD}]	{
$$
\xymatrix@=6mm
{ X_{111} & &*=<25pt,10pt>[F-:<3pt>]{n_1}
\ar@/_/@{-}[ldd] \ar@/^/@{--}[dr]\\ 
X_{121}  & & &*=<25pt,10pt>[F-:<3pt>]{n_2} 
\ar@/_/@{-}[dll]\ar@{--}[dd] 
\\
X_{211}& *=<25pt,10pt>[F-:<3pt>]{n_3}
\ar@{-}[d] \ar@/^/@{--}[drr]  \\
&*=<25pt,10pt>[F]{1} &&*=<25pt,10pt>[F]{0}}
$$	
}	
\caption{Decision diagrams for the query \textit{eruption} of Example  \ref{stromboli}. \ref{MDD}  the MDD representing the set of explanations encoded by function (\ref{Feruption}), built with multi-valued variables; \ref{BDD} the corresponding BDD representing the set of explanations encoded by function (\ref{Feruption_bdd}), built with binary variables.}
\label{dd}
\end{figure}

An MDD can be used to represent $f_Q(\mathbf X)$ and,
since MDDs split paths on the basis of the values of a variable, the branches are mutually disjoint so the dynamic programming algorithm of \cite{DBLP:conf/ijcai/RaedtKT07} can be applied for computing the probability of $Q$.
For example,
the reduced MDD corresponding to the query $Q=eruption$ from Example \ref{stromboli} is shown in Figure \ref{MDD}. The labels on the edges represent the values of the variable associated with the source node.

Most packages for the manipulation of  decision diagrams are however restricted to work on Binary Decision Diagrams (BDDs), i.e., decision diagrams where all the variables are Boolean.
These packages offer Boolean operators among BDDs and apply simplification rules to the result of the operations in order to reduce as much as possible the BDD size.

A node $n$ in a BDD has two children: one corresponding to the 1 value of the variable associated with $n$, indicated with $child_1(n)$, and one corresponding the 0 value of the variable, indicated with $child_0(n)$. When drawing BDDs, rather than using edge labels, the 0\--branch - the one going to  $child_0(n)$ -  is distinguished from the 1-branch by drawing it with a dashed line.

To work on MDDs with a BDD package we must represent multi-valued variables by means of binary variables. Various options are possible, we found that the following 
provides a good performance \cite{DBLP:conf/aaai/SangBK05,DeR-NIPS08}: for a multi\--valued variable $X_{ij}$, corresponding to the ground clause $C_{i}\theta_j$, having $n_i$ values, we use $n_i-1$ Boolean variables $X_{ij1},\ldots,X_{ijn_i-1}$ and we represent the equation $X_{ij}=k$ for $k=1,\ldots n_i-1$ by means of the conjunction  $\overline{X_{ij1}}\wedge\ldots \wedge \overline{X_{ijk-1}}\wedge X_{ijk}$, and the equation $X_{ij}=n_i$ by means of the conjunction  $\overline{X_{ij1}}\wedge\ldots \wedge\overline{X_{ijn_i-1}}$. \\

 According to the above transformation, $X_{11}$ and $X_{12}$ are 3-valued variables and each one is converted into two Boolean variables  ($X_{111}$ and $X_{112}$ for the former, $X_{121}$ and $X_{122}$ for the latter); $X_{21}$ is a 2-valued variable and is converted into the Boolean variable $X_{211}$.
The set of explanations $E(eruption)=\{\kappa_1, \kappa_2\}$ can be now encoded by the equivalent function
\begin{equation}
\label{Feruption_bdd}
f'_{eruption}(\mathbf{X})=(X_{111}  \wedge X_{211}) \vee (X_{121}  \wedge X_{211})
\end{equation}
with the first disjunct representing $\kappa_1$ and the second disjunct $\kappa_2$.
The BDD encoding of $f'_{eruption}(\mathbf{X})$ is shown in Figure \ref{BDD} and corresponds to the MDD of Figure \ref{MDD}.
A value 1 for the Boolean variables $X_{111}$ and $X_{121}$ means that, for the ground clauses $C_1\theta_1$ and $C_1\theta_2$, the head atom $h_{11} = eruption$ is chosen and the 1-branch from nodes $n_1$ and $n_2$ must be followed, regardless of the other variables for $C_1$ ($X_{112}$, $X_{122}$) that are in fact omitted from the diagram.

BDDs obtained in this way can be used as well for computing the probability of queries by associating with every Boolean variable $X_{ijk}$  a parameter $\pi_{ik}$ that represents $P(X_{ijk}=1)$. 
The parameters are obtained from those of multi-valued variables in this way: 
\begin{eqnarray*}
\pi_{i1}&=&\Pi_{i1}\\
\ldots \\
\pi_{ik}&=&\frac{\Pi_{ik}}{\prod_{j=1}^{k-1}(1-\pi_{ij})}\\
\ldots
\end{eqnarray*}
up to $k=n_i-1$.

The use of BDDs for probabilistic logic programming inference is related to their use for performing inference in Bayesian networks. \citeANP{DBLP:conf/ijcai/MinatoSS07}  \citeyear{DBLP:conf/ijcai/MinatoSS07}  presented a method for compiling BNs into exponentially-sized Multi-Linear Functions using a compact Zero-suppressed BDD representation.  \citeANP{DBLP:conf/ausai/IshihataSM11} \citeyear{DBLP:conf/ausai/IshihataSM11} compile a BN with multiple evidence sets into a single Shared BDD, which shares common sub-graphs in multiple BDDs. \citeANP{DBLP:conf/ecai/Darwiche04}  \citeyear{DBLP:conf/ecai/Darwiche04} described an algorithm for compiling propositional formulas in conjunctive normal form into Deterministic Decomposable Negation Normal Form (d-DNNF) - a tractable logical form for model
counting in polynomial time - with techniques from the Ordered BDD literature.

\section{\emalgorithm}
\label{emblem}

\emalgorithm{} \cite{BelRig13-IDA-IJ} learns LPAD parameters by using an Expectation Maximization algorithm where the expectations are computed directly on BDDs. It is based on the algorithms proposed in \cite{Ish-ILP2008,DBLP:conf/pkdd/ThonLR08,Ish-TR08,DBLP:conf/ijcai/InoueSIKN09}. While using a similar approach with respect to LFI-ProbLog \cite{DBLP:conf/pkdd/GutmannTR11},
 \emalgorithm{} is more targeted at discriminative learning \cite{BelRig12-IA-IJ}, so we chose it for parameter learning in SLIPCOVER. In particular, \emalgorithm{} and LFI-Problog differ in the construction of BDDs: LFI-ProbLog builds a BDD for
a whole partial interpretation while EMBLEM for single ground atoms for the specified target predicate(s), the one(s)
for which we are interested in good predictions. Moreover LFI-ProbLog treats missing nodes as if they were there and
updates the counts accordingly, while we compute the contributions of deleted paths with the $\varsigma$ table.

The typical input for \emalgorithm\ is a set of \textit{target} predicates, a set of \emph{mega-examples} and a theory. The mega-examples  are sets of ground facts describing a portion of the domain of interest  and must contain also negative facts for target predicates, expressed as $neg(atom)$. 
Among the predicates describing the domain, the user has to indicate which are target: the facts for these predicates in the mega-examples will form the queries $Q$ for which the BDDs are built, encoding the disjunction of their explanations  (cf. Figure \ref{BDD}).
The input theory is an LPAD.

EMBLEM applies an  Expectation Maximization (EM) algorithm where the expectations in the E-step are computed directly on the BDDs built for the target facts.
The  predicates can be treated as closed\--world or open\--world. In the first case, the body of clauses is resolved only with facts in the mega-example. In the second case, the body of clauses is resolved both with facts  and with clauses in the theory. If the latter option is set and the theory is cyclic (composed of recursive clauses), EMBLEM uses a depth bound on SLD-derivations to avoid going into infinite loops, as proposed by \cite{gutmann:tr2010-1}, with $D$ the value of the bound.

After building the BDDs, \emalgorithm\ starts the EM cycle, in which the steps of Expectation and Maximization are repeated until the LL of the examples reaches a local maximum or a maximum number of steps ($NEM$) is executed. %
 \emalgorithm\  is shown in Algorithm  \ref{EM}: it consists of a cycle where the functions \textsc{Expectation} and \textsc{Maximization} are repeatedly called; function \textsc{Expectation} returns the LL of the data that is used in the stopping criterion. \emalgorithm{} stops when the difference between the LL of the current and  the previous iteration drops below a threshold $\epsilon$ or when this difference is below a fraction $\delta$ of the current LL.

\begin{algorithm}[ht]\begin{scriptsize}
\caption{Function EMBLEM} \label{EM}
\begin{algorithmic}[1]
\Function{EMBLEM}{$Theory,D,NEM,\epsilon,\delta$}
\State Build $BDDs$ by SLD derivations with depth bound $D$
\State $LL=-\mathit{inf}$
\State $N=0$
\Repeat   \Comment{Start of EM cycle}
\State $LL_0=LL$
\State $LL=$ \Call{Expectation}{$BDDs$}
\State \Call{Maximization}{}
\State $N=N+1$
\Until{$LL-LL_0 < \epsilon \vee LL-LL_0 < - LL \cdot \delta\vee N>NEM$} 
\State Update parameters of $Theory$ 
\State return $(LL,Theory)$
\EndFunction
\end{algorithmic}
\end{scriptsize}
\end{algorithm}

The Expectation phase (see Algorithm \ref{exp}) takes as input a list of BDDs, one for each target fact Q, and computes the expectations $\mathbf{E}[X_{ijk}=x|Q]$ for all $C_i$s, $k=1,\ldots,n_i-1$, $j\in g(i):=\{j|\theta_j$  is  a  substitution  grounding $C_i\}$ and $x\in \{0,1\}$. 
$\mathbf{E}[X_{ijk}=x|Q]$  is given by
$$\mathbf{E}[X_{ijk}=x|Q]=P(X_{ijk}=x|Q)\cdot 1+P(X_{ijk}=(1-x)|Q)\cdot 0=P(X_{ijk}=x|Q)$$
From $\mathbf{E}[X_{ijk}=x|Q]$ we can compute the expectations
$\mathbf{E}[c_{ik0}|Q]$ and $\mathbf{E}[c_{ik1}|Q]$
where $c_{ikx}$ is the number of times a Boolean variable $X_{ijk}$ takes on value $x$ for $x\in \{0,1\}$ and for all $j\in g(i)$. 
$\mathbf{E}[c_{ikx}|Q]$ is given by 
$$\mathbf{E}[c_{ikx}|Q]=\sum_{j\in g(i)}P(X_{ijk}=x|Q)$$
In this way the variables $X_{ijk}$ with the same $j$ are independent and identically distributed. Finally, the  expectations $\mathbf{E}[c_{ik0}]$ and $\mathbf{E}[c_{ik1}]$ of the counts over all queries are computed as
$$\mathbf{E}[c_{ikx}]=\sum_{Q}\mathbf{E}[c_{ikx}|Q]$$
\algrenewcommand\alglinenumber[1]{\scriptsize #1:}
\begin{algorithm}[ht]\begin{scriptsize}
\caption{Function Expectation} \label{exp}
\begin{algorithmic}[1]
\Function{Expectation}{$BDDs$}
\State $LL=0$
\ForAll{$BDD\in BDDs$}
\ForAll{$i\in Rules$}
\For{$k=1$ to $n_i-1$}
\State $\eta^0(i,k)=0$;\ \ $\eta^1(i,k)=0$
\EndFor
\EndFor
\ForAll{variables X}
\State $\varsigma(X)=0$
\EndFor
\State \Call{GetForward}{$root(BDD)$}
\State $Prob$=\Call{GetBackward}{$root(BDD)$}
\State $T=0$
\For{$l=1$ to $levels(BDD)$}
\State Let $X_{ijk}$ be the variable associated with level $l$
\State $T=T+\varsigma(l)$
\State $\eta^0(i,k)=\eta^0(i,k)+T\times(1-\pi_{ik})$
\State $\eta^1(i,k)=\eta^1(i,k)+T\times\pi_{ik}$
\EndFor
\ForAll{$i\in Rules$}
\For{$k=1$ to $n_i-1$}
\State $\mathbf{E}[c_{ik0}]=\mathbf{E}[c_{ik0}]+\eta^0(i,k)/Prob$
\State $\mathbf{E}[c_{ik1}]=\mathbf{E}[c_{ik1}]+\eta^1(i,k)/Prob$
\EndFor
\EndFor
\State $LL=LL+\log(Prob)$
\EndFor
\State return $LL$
\EndFunction
\end{algorithmic}
\end{scriptsize}
\end{algorithm}

\begin{algorithm}[ht]\begin{scriptsize}
\caption{Procedure Maximization} \label{max}
\begin{algorithmic}[1]
\Procedure{Maximization}{}
\ForAll{$i\in Rules$}
\For{$k=1$ to $n_i-1$}
\State $\pi_{ik}=\frac{\mathbf{E}[c_{ik1}]}{\mathbf{E}[c_{ik0}]+\mathbf{E}[c_{ik1}]}$
\EndFor
\EndFor
\EndProcedure
\end{algorithmic}
\end{scriptsize}
\end{algorithm}

$P(X_{ijk}=x|Q)$ is given by $\frac{P(X_{ijk}=x,Q)}{P(Q)}$ , where
\begin{eqnarray*}
P(X_{ijk}=x,Q)
&=&
\sum_{l_\sigma\in L_T:l_\sigma\models Q}P(Q,X_{ijk}=x,\sigma)\\
&=&\sum_{l_\sigma\in L_T:l_\sigma\models Q}P(Q|\sigma)P(X_{ijk}=x|\sigma)P(\sigma)\\
&=&\sum_{l_\sigma\in L_T:l_\sigma\models Q}P(X_{ijk}=x|\sigma)P(\sigma)%
\end{eqnarray*}
Now suppose that only the merge rule is applied when building the BDD, fusing together identical sub-diagrams.  %
The result, that we call Complete Binary Decision Diagram (CBDD), is such that every path contains a node at every level.

Since there is a one to one correspondence between the  instances where $Q$ is true and the paths to a 1 leaf in a CBDD,
$$P(X_{ijk}=x,Q)=\sum_{\rho\in R(Q)}P(X_{ijk}=x|\rho)\prod_{d\in \rho}\pi(d)$$
where $\rho$ is a path and, if $\sigma$ corresponds to $\rho$, then $P(X_{ijk}=x|\sigma)$=$P(X_{ijk}=x|\rho)$. $R(Q)$ is the set of paths in the CBDD for query $Q$ that lead to a 1 leaf, $d$ is an edge of $\rho$ and $\pi(d)$ is the probability associated with the edge: if $d$ is the $1$\--branch from a node associated with a variable $X_{ijk}$, then $\pi(d)=\pi_{ik}$,
if $d$ is the $0$\--branch, then $\pi(d)=1-\pi_{ik}$.

Given a path $\rho\in R(Q)$, 
$P(X_{ijk}=x|\rho)=1$ if $\rho$ contains an $x$-branch from a node associated with variable $X_{ijk}$ and 0 otherwise, so $P(X_{ijk}=x,Q)$ can be further expanded as
\begin{eqnarray*}
P(X_{ijk}=x,Q)&=&\sum_{\rho\in R(Q)\wedge(X_{ijk}=x) \in \rho}\prod_{d\in \rho}\pi(d)\\
\end{eqnarray*}
where $(X_{ijk}=x) \in \rho$ means that $\rho$ contains an $x$\--branch from the node associated with 
$X_{ijk}$. We can then write
\begin{eqnarray*}
P(X_{ijk}=x,Q)
&=&\sum_{n \in N(Q)\wedge v(n)=X_{ijk} \wedge \rho_n \in R_n(Q)\wedge\rho^n \in R^n(Q,x)}\prod_{d\in \rho^n}\pi(d)\prod_{d\in \rho_n}\pi(d)\\
\end{eqnarray*}
where $N(Q)$ is the set of BDD nodes, $v(n)$ is the variable associated with node $n$, $R_n(Q)$ is the set of  paths from the root to $n$ and $R^n(Q,x)$ is the set of paths from $n$ to the 1 leaf through its $x$\--child. So
\begin{eqnarray*}
P(X_{ijk}=x,Q)
&=&\sum_{n \in N(Q)\wedge v(n)=X_{ijk}}\sum_{\rho_n \in R_n(Q)}\sum_{\rho^n \in R^n(Q,x)}\prod_{d\in \rho^n}\pi(d)\prod_{d\in \rho_n}\pi(d)\\
&=&\sum_{n \in N(Q)\wedge v(n)=X_{ijk}}\sum_{\rho_n \in R_n(Q)}\prod_{d\in \rho_n}\pi(d)\sum_{\rho^n \in R^n(Q,x)}\prod_{d\in \rho^n}\pi(d)\\ 
&=&\sum_{n\in N(Q)\wedge v(n)=X_{ijk}}F(n)B(child_x(n))\pi_{ikx}
\end{eqnarray*}
where $\pi_{ikx}$ is $\pi_{ik}$ if $x=1$ and  $(1-\pi_{ik})$ if $x=0$, 
$F(n)=\sum_{\rho_n \in R_n(Q)}\prod_{d\in \rho_n}\pi(d)$
is the \emph{forward probability} \cite{Ish-TR08}, the probability mass of the paths from the root to $n$, and 
$B(n)=\sum_{\rho^n \in R^n(Q)}\prod_{d\in \rho^n}\pi(d)$
is the \emph{backward probability} \cite{Ish-TR08}, the probability mass of paths from $n$ to the 1 leaf. Here $R^n(Q)$ is the set of paths from $n$ to the 1 leaf. If $root$ is the root of a tree for a query $Q$ then $B(root)=P(Q)$,
which is needed to compute $P(X_{ijk}=x|Q)$.
The expression $F(n)B(child_x(n))\pi_{ikx}$ represents the sum of the probabilities of all the paths passing through the $x$\--edge of node $n$. By indicating with $e^x(n)$ such an expression we get

\begin{equation}\label{pe}
P(X_{ijk}=x,Q)=\sum_{n \in N(Q),v(n)=X_{ijk}}e^x(n)
\end{equation}
Computing the forward probability 
and the backward probability 
of BDDs'  nodes requires two traversals of the graph, so the cost is linear in the number of nodes.
The counts are stored in  variables $\eta^{x}(i,k)$ for $x\in \{0,1\}$. In the end  $\eta^{x}(i,k)$ contains
$$\sum_{j\in g(i)}P(X_{ijk}=x,Q)$$
\begin{figure}[ht]
\framebox[12.1cm]{
\begin{minipage}{12.1cm}
	{
\scriptsize	
$$\xymatrix@R=4mm@C=2mm
{ X_{111} & &*=<52pt,15pt>[F-:<3pt>]{n_1\ \parbox{36pt}{$F=1$\\
$B=0.588$}}
\ar@/_/@{-}[ldd]^{0.6} \ar@/^/@{--}[dr]_{0.4}\\ 
X_{121}  & & &*=<52pt,15pt>[F-:<3pt>]{n_2\ \parbox{36pt}{F=0.4\\B=0.42}} 
\ar@/_/@{-}[dll]^{0.6}\ar@{--}[dd] _{0.4}
\\
X_{211}& *=<52pt,15pt>[F-:<3pt>]{n_3\ \parbox{35pt}{F=0.84\\B=0.7}}
\ar@{-}[d]^{0.7} \ar@/^/@{--}[drr]^{0.3}  \\
&*=<52pt,15pt>[F]{1} &&*=<52pt,15pt>[F]{0}}
$$
}%
\end{minipage}
}%
\caption{Forward and backward probabilities for Example 3. $F$ indicates the forward probability and $B$ the backward probability of each node.}
\label{FBProb}
\end{figure}
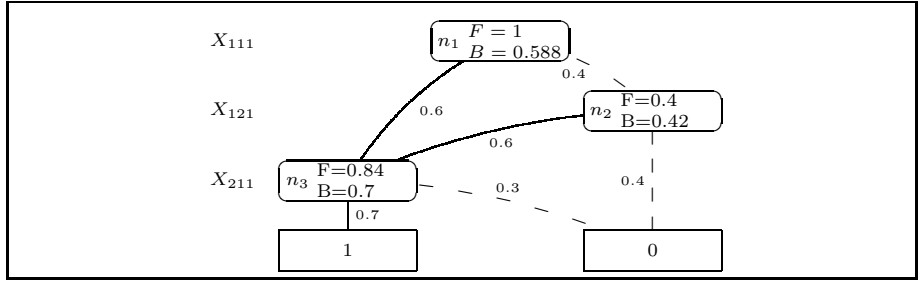
Formula (\ref{pe}) is correct if, when building the BDD, no node has been deleted, i.e., if a node for every variable appears on each path. If this is not the case, the contribution of deleted paths must be taken into account. This is done in the algorithm of \cite{Ish-ILP2008} by keeping an array $\varsigma$ with an entry for every level $l$ that stores an algebraic sum of  $e^x(n)$.

In the Maximization phase (see Algorithm \ref{max}), the  $\pi_{ik}$ parameters are computed for all rules $C_i$ and $k=1,\ldots,n_i-1$ as
$$\pi_{ik}=\frac{\mathbf{E}[c_{ik1}]}{\mathbf{E}[c_{ik0}]+\mathbf{E}[c_{ik1}]}$$
for the next EM iteration.

Suppose you have the program of Example \ref{stromboli} and you have the single example $Q=eruption$. The BDD of Figure \ref{BDD}  is built and passed to \textsc{Expectation} in the form of a pointer to its root node $n_1$. The resulting forward and backward probabilities are shown in Figure \ref{FBProb}.

\section{SLIPCOVER}
\label{slipcover}
SLIPCOVER learns an LPAD by first identifying good candidate
clauses and then by searching for a theory guided by the LL of the data.
As EMBLEM, it takes as input a set of mega-examples and an indication of which predicates are target, i.e., those for which we want to optimize the predictions of the final theory. The mega-examples must contain positive and negative examples for all predicates that may appear in the head of clauses, either target or non-target (background predicates).

\subsection{The language bias}
\label{lb}
The search over the space of clauses to identify the candidate ones is performed according to a language bias expressed by means of \textit{mode} declarations.
Following \cite{DBLP:journals/ngc/Muggleton95}, 
a mode declaration $m$ is either a head declaration $modeh(r , s)$ or a body declaration $modeb(r , s)$, where $s$, the \emph{schema}, is a ground
literal and $r$ is an integer called the \textit{recall}.
A schema is a template for literals in the head or body of a  clause and can contain special placemarker terms of the form \verb|\#type|, \verb|+type| and \verb|-type|, which stand, respectively, for ground terms, input variables and output
variables of  a type. An input variable in a body literal of a clause must be either
an input variable in the head or an output variable in a preceding body literal  in the clause.
If $M$ is a set of mode declarations, $L( M )$ is the \emph{language of} $M$, i.e. the set of clauses $\{C = h_1;\ldots; h_n \ {:\!-}\   b_1 , \ldots, b_m\}$ such that the head atoms $h_i$ (resp. body literals $b_i$) are obtained from some head (resp. body) declaration in $M$
by replacing all \#    placemarkers with ground terms and all + (resp. -) placemarkers with input (resp. output) variables. 
We extend this type of mode declarations with placemarker terms of the form -\# which are treated as \# when defining $L(M)$ but differ in the creation of the bottom clauses, see subsection \ref{ib}. These mode declarations are used also by SLIPCASE.

We extended the mode declarations with respect to SLIPCASE by allowing  head declarations of the form $modeh(r,[s_1,\ldots,s_n],[a_1,\ldots,a_n],[P_1/Ar_1,\ldots,P_k/Ar_k])$. These are used to generate clauses with more than two head atoms. $s_1,\ldots,s_n$ are schemas,  $a_1,\ldots,a_n$ are atoms such that $a_i$ is obtained from $s_i$ by replacing placemarkers with variables, $P_i/Ar_i$ are the predicates admitted in the body. $a_1,\ldots,a_n$ are used to indicate which variables should be shared by the atoms in the head.

Examples of mode declarations can be found in subsection \ref{example_advisedby}.

\subsection{Description of the algorithm}
The main function is shown by Algorithm \ref{SEM}: after the search in the space of clauses, encoded in lines 2 - 27,  SLIPCOVER performs a greedy  search in the space of theories, described in lines 28 - 39. 

The first  phase aims at searching \emph{in the space of clauses} in order to find a set of promising ones (in terms of LL of the data), that will be employed in the subsequent greedy search phase. By starting from promising clauses, the greedy search is able to generate good final theories.
The search in the space of clauses is split in turn in two steps: (1) the construction of a set of beams containing the bottom clauses (function \textsc{InitialBeams} at line 2 of Algorithm \ref{SEM}), and (2) a beam search over each of these beams to refine the bottom clauses (function \textsc{ClauseRefinements} at line 11). The overall output of this search phase is represented by two lists of refined promising clauses: $TC$ for target predicates and $BC$ for background predicates. The clauses are inserted in $TC$ if  a target predicate appears in their head, otherwise in $BC$. The lists are sorted in decreasing LL.

\algrenewcommand\alglinenumber[1]{\scriptsize #1:}
\begin{algorithm}[ht]\begin{scriptsize}
\caption{Function \textsc{SLIPCOVER}} \label{SEM}
\begin{algorithmic}[1]
\Function{SLIPCOVER}{$NInt,NS,NA,NI,NV,NB,NTC,NBC,D,NEM,\epsilon,\delta$}
\State $IBs=$\Call{InitialBeams}{$NInt,NS,NA$}      \Comment{Clause search}
 \State $TC\gets []$
 \State $BC\gets []$
\ForAll{$(PredSpec,Beam) \in IBs$}
 \State $Steps\gets 1$
 \State $NewBeam\gets []$
 \Repeat   
    \While{$Beam$  is not empty}
	\State Remove the first couple $((Cl,Literals),LL)$ from $Beam$   \Comment{Remove the first  clause}
	\State $Refs\gets $\Call{ClauseRefinements}{$(Cl,Literals),NV$} \Comment{Find all refinements $Refs$ of $(Cl,Literals)$ with at most $NV$ variables}
	\ForAll{$(Cl',Literals') \in \Refs$}
		\State $(LL'',\{Cl''\})\gets $\Call{EMBLEM}{$\{Cl'\},D,NEM,\epsilon,\delta$}
		\State $NewBeam\gets $\Call{Insert}{$(Cl'',Literals'),LL'',NewBeam,NB$}
		\If{$Cl''$ is range restricted}
		  \If{$Cl''$ has a target predicate in the head}
		      \State $TC\gets $\Call{Insert}{$(Cl'',Literals'),LL'',TC,NTC$}
            \Else	    
		      \State $BC\gets $\Call{Insert}{$(Cl'',Literals'),LL'',BC,NBC$}
		    \EndIf
		\EndIf
	\EndFor
	\EndWhile
	\State $Beam\gets NewBeam$
	\State $Steps\gets Steps+1$
  \Until{$Steps > NI$}
\EndFor
\State $Th\gets \emptyset$, $ThLL\gets -\infty$    \Comment{Theory search}
\Repeat
	\State Remove the first couple $(Cl,LL)$ from $TC$
		\State $(LL',Th')\gets $\Call{EMBLEM}{$Th\cup\{Cl\},D,NEM,\epsilon,\delta$}
		\If{$LL'> ThLL$}
			\State $Th\gets Th'$, $ThLL\gets LL'$
		\EndIf    
\Until{$TC$ is empty}
\State $Th\gets Th\bigcup_{(Cl,LL)\in BC}\{Cl\}$
	\State $(LL,Th)\gets $\Call{EMBLEM}{$Th,D,NEM,\epsilon,\delta$}
\State return $Th$
\EndFunction
\end{algorithmic}
\end{scriptsize}
\end{algorithm}

The second phase is a greedy search \emph{in the space of theories} starting with an empty theory $Th$ with the lowest value of LL (line 28 of Algorithm \ref{SEM}). Then one target clause $Cl$ at a time  is added  from the list $TC$. 
After each addition, \emalgorithm{} is run on the extended theory $Th \cup \{Cl\}$ and the log likelihood $LL'$ of the data is computed as the score of the resulting theory $Th'$.  If  $LL'$ is better than the current best, the clause is kept in the theory, otherwise it is discarded (cf. lines 31-33). This is done for each clause in $TC$.

Finally, SLIPCOVER adds all the (background) clauses in the list $BC$ to the theory composed only of target clauses (cf. line 36) and performs parameter learning on the resulting theory (cf. line 37). The clauses that are never used to derive the examples will get a value of 0 for the parameters of the atoms in their head and will be removed in a post processing phase.

In the following we provide a detailed description of the two support functions for the first phase, the \emph{search in the space of clauses}.

\subsubsection{Function \textsc{InitialBeams}}
\label{ib}
Algorithm \ref{initialbeam} shows how the initial set of beams $IB$, one for each predicate $P$ (with arity $Ar$) appearing in a modeh declaration, is generated by SLIPCOVER by building a set of \emph{bottom clauses} as in Progol \cite{DBLP:journals/ngc/Muggleton95}, by means of the predefined language bias (cf. subsection \ref{lb}). The algorithm outputs the initial clauses that will be subsequently refined by Function \textsc{ClauseRefinements}.

In order to generate a bottom clause for a mode declaration $modeh(r,s)$ specified in the language bias, an input mega-example $I$ is selected and an answer $h$ for the goal $schema(s)$ is selected, where $schema(s)$ denotes the literal obtained from $s$ by replacing all placemarkers with distinct variables $X_1 ,\ldots, X_n$ (cf. lines 5-9 of Algorithm \ref{initialbeam}). Both the mega-example and the atom $h$ are randomly sampled with replacement, the former from the available set of training mega-examples and the latter from the set of all answers found for the goal $schema(s)$.

Then $h$ is saturated with body literals using  Progol's  saturation method, encoded in Function \textsc{Saturation} shown in Algorithm \ref{progol}.
This method is a deductive procedure used to find atoms related to $h$. The terms in $h$ are  used to initialize a growing set of input terms $InTerms$:  these are the terms corresponding to + placemarkers in $s$. 
Then each body declaration $m$ is considered in turn. The terms from $InTerms$ are substituted into the + placemarkers of $m$ to generate a set $Q$ of goals. Each goal is then executed against the database and up to $r$ (the recall) successful ground instances (or all if $r=\star$) are added to the body of the clause; only positive examples are considered to solve the goal. Any  term corresponding to a - or -\# placemarker in $m$  is inserted in $InTerms$ if it is not already present. %
This cycle is repeated for an user-defined number $NS$ of times.

The resulting ground clause $BC = h \ {:\!-}\   b_1,\ldots,b_m$ is then processed to obtain a program clause by replacing each term in a + or - placemarker with a variable, using the same variable for identical terms. Terms corresponding to \# or -\# placemarkers are instead kept in the clause. 
The initial beam $Beam$ associated with predicate $P/Ar$ of $h$ will contain the clause with empty body $h:0.5.$ for each bottom clause $h\ {:\!-}\   b_1,\ldots,b_m$ (cf. lines 10-11 of Algorithm \ref{initialbeam}). 
This process is repeated for a number $NInt$ of input mega-examples and a number $NA$ of answers, thus obtaining $NInt\cdot NA$ bottom clauses.

The generation of a bottom clause for a mode declaration 

$m=modeh(r,[s_1,\ldots,s_n],[a_1,\ldots,a_n],[P_1/Ar_1,\ldots,P_k/Ar_k])$

\noindent 
is the same except for the fact that the goal to call is composed of more than one atom. In order to build the head, the goal $a_1,\ldots,a_n$ is called and $NA$ answers that ground all $a_i$s are kept (cf. lines 15-19). From these, the set of input terms $InTerms$ is built and body literals are found by the Function \textsc{Saturation} (cf. line 20 of Algorithm \ref{initialbeam}) as above. The resulting bottom clauses then have the form
$a_1 \ ;\ \ldots \ ; \ a_n \ {:\!-}\   b_1,\ldots,b_m$ and the initial beam $Beam$ will contain  clauses with an empty body of the form $a_1:\frac{1}{n+1}\ ;\ \ldots \ ; \ a_n:\frac{1}{n+1}.$ (cf. line 21 of Algorithm \ref{initialbeam}) \\
Finally, the set of the beams for each predicate $P$ is returned to the Function \textsc{Slipcover}.

\begin{algorithm}[ht]\begin{scriptsize}
\caption{Function \textsc{InitialBeams}} \label{initialbeam}
\begin{algorithmic}[1]
\Function{InitialBeams}{$NInt,NS,NA$}
\State $IB\gets\emptyset$
\ForAll{predicates $P/Ar$}
  \State $Beam\gets []$
  \ForAll{modeh declarations $modeh(r,s)$ with $P/Ar$ predicate of $s$}
      \For{$i=1 \to NInt$}
	    \State Select randomly a mega-example $I$
	    \For{$j=1 \to NA$}
	      \State Select randomly an atom $h$ from $I$ matching $schema(s)$
	       \State Bottom clause $BC\gets $\Call{Saturation}{$h,r,NS$}, let $BC$ be $Head\ {:\!-}\   Body $
	      \State $Beam\gets  [((h:0.5,Body),-\infty)|Beam]$
         \EndFor
       \EndFor
  \EndFor
  \ForAll{modeh declarations $modeh(r,[s_1,\ldots,s_n],[a_1,\ldots,a_n],PL)$ with $P/Ar$ in $PL$ appearing in $s_1,\ldots,s_n$}
      \For{$i=1 \to NInt$}
	    \State Select randomly a mega-example $I$
	    \For{$j=1 \to NA$}
	      \State Select randomly a set of  atoms $h_1,\ldots,h_n$ from $I$ matching $a_1,\ldots,a_n$ 
	       \State Bottom clause $BC\gets $\Call{Saturation}{$(h_1,\ldots,h_n),r,NS$}, let $BC$ be $Head\ {:\!-}\   Body $
	      \State $Beam\gets  [((a_1:\frac{1}{n+1}\ ;\ \ldots \ ; \ a_n:\frac{1}{n+1},Body),-\infty)|Beam]$
         \EndFor
       \EndFor
  \EndFor
  \State $IB\gets IB\cup\{(P/Ar,Beam)\}$
\EndFor
\State return $IB$
\EndFunction
\end{algorithmic}
\end{scriptsize}
\end{algorithm}

\algrenewcommand\alglinenumber[1]{\scriptsize #1:}
\begin{algorithm}[ht]\begin{scriptsize}
\caption{Function \textsc{Saturation}} \label{progol}
\begin{algorithmic}[1]
\Function{Saturation}{$Head,r,NS$}
\State $InTerms=\emptyset$, 
\State $BC=\emptyset$   \Comment{$BC$: bottom clause}
\ForAll{arguments $t$ of $Head$}
				\If{$t$ corresponds to a $+type$}
					\State add $t$ to $InTerms$
				\EndIf
\EndFor
\State Let $BC$'s head be $Head$
\Repeat
 \State $Steps \gets 1$
\ForAll{modeb declarations $modeb(r,s)$}
	\ForAll {possible subs. $\sigma$ of variables corresponding to $+type$ in $schema(s)$ by terms from $InTerms$}
		\For{$j=1 \to r$}
			\If{goal $b=schema(s)$ succeeds with answer substitution $\sigma'$}
				\ForAll{$v/t \in \sigma$ and $\sigma'$}
		        \If{$v$ corresponds to a $-type$ or $-\#type$}
							\State add $t$ to the set $InTerms$ if not already present
						\EndIf
				 \EndFor
			 \State Add $b$ to $BC$'s body
			\EndIf
		\EndFor	
 \EndFor
\EndFor
 \State $Steps \gets Steps+1$
\Until $Steps > NS$
\State Replace constants with variables in $BC$, using the same variable for equal terms
\State return $BC$
\EndFunction
\end{algorithmic}
\end{scriptsize}
\end{algorithm}

\subsubsection{Beam Search with Clause Refinements}
\label{beam_clauseref}

After having built the initial bottom clauses gathered in beams,
a cycle on every predicate, either target or background, is performed (line 5 of Algorithm \ref{SEM}): in each iteration, SLIPCOVER runs a beam search in the space of clauses for the predicate (line 9).

For each bottom clause $Cl$  with  $Literals$ admissible in the body, Function\\ \textsc{ClauseRefinements}, shown in Algorithm \ref{CRef}, computes refinements by adding a literal from $Literals$ to the body or deleting an atom from the head  in the case of multiple-head bottom clauses with a number of disjuncts (including the \textit{null} atom) greater than 2.
Furthermore, the refinements must respect the input-output modes of the bias declarations, must be connected (i.e., each body literal must share a variable with the head or a previous body literal) and their number of variables must not exceed a user-defined number $NV$.
The couple ($Cl'$, $Literals'$) indicates a refined clause $Cl'$ together with the new set $Literals'$ of literals allowed in the body of $Cl'$; the tuple ($Cl_h'$, $Literals$) indicates a specialized clause $Cl'$ where one disjunct in its head has been removed.

At line 13 of Algorithm \ref{SEM},  parameter learning  is executed for a theory composed of the single refined clause. For each goal for the current predicate, \emalgorithm{} builds the BDD encoding its explanations by deriving them from the single-clause theory together with the facts in the mega-examples; derivations exceeding the depth limit $D$ are cut. Then the parameters and the LL of the data are computed by the EM algorithm; LL is used as score of the updated clause $(Cl'',Literals')$.

This clause is then inserted into a list of promising clauses: %
in $TC$ if  a target predicate appears in its head, otherwise in $BC$. The insertion is in order of decreasing LL. 
 If the clause is not range restricted, i.e., if some of the variables in the head do not appear in a positive literal in the body, then it is not inserted  in $TC$ nor in $BC$. 
These lists have a maximum size: if an insertion increases the size over the maximum, the last element is removed. 
In Algorithm \ref{SEM}, the Function \textsc{Insert($I,Score,List,N$)} is used to insert in order a clause $I$ with score  $Score$ in a  $List$ with at most $N$ elements. Beam search is repeated until the beam becomes empty or a maximum number $NI$ of iterations is reached.

The separate search for clauses has similarity with the covering loop of  ILP systems such as Aleph and Progol. Differently from the ILP case, however, the test of an example requires the computation of all its explanations, while in ILP the search stops at the first matching clause.
The only interaction among clauses in probabilistic logic programming happens if the clauses are recursive. If not, then adding clauses to a theory only adds explanations for the example - increasing its probability - so clauses can be added individually to the theory. If the clauses are recursive, the examples for the head predicates are used to resolve literals in the body, thus the test of examples on individual clauses approximates the case of the test on a complete theory. As will be shown by the experiments, this approximation is often sufficient for identifying good clauses.

\algrenewcommand\alglinenumber[1]{\scriptsize #1:}
\begin{algorithm}[ht]\begin{scriptsize}
\caption{Function \textsc{ClauseRefinements}} \label{CRef}
\begin{algorithmic}[1]
\Function{ClauseRefinements}{$(Cl,Literals), NV$} %
\State $Refs = \emptyset$, $Nvar=0$;  \Comment{Nvar:number of different variables in a clause}
	\ForAll{$b\in Literals$} 
	\State $Literals'\gets Literals \setminus \{b\}$ 
	  \State Add $b$ to $Cl$ body obtaining $Cl'$
	  \State $Nvar \gets$ number of $Cl'$  variables
			\If{$Cl'$ is connected $\wedge$ $Nvar < NV$}
				\State $Refs\gets Refs\cup \{(Cl',Literals')\}$				
		   \EndIf
		\EndFor
		  \If{$Cl$ is a multiple-head clause} \Comment{It has 3 or more disjuncts including the \textit{null} atom}
		  	\State Remove one  atom from $Cl$ head obtaining $Cl_h'$ \Comment{Not the \textit{null} atom}
		  	\State Adjust the probabilities on the remaining head atoms
		  	\State $Refs\gets Refs\cup \{(Cl_h',Literals')\}$				
		  \EndIf

\State return $Refs$
\EndFunction
\end{algorithmic}
\end{scriptsize}
\end{algorithm}

\subsection{Execution example}
\label{example_advisedby}
We now show an example of execution for the UW-CSE dataset that is used in the experiments discussed in Section \ref{experiments}. UW-CSE describes the Computer Science department of the University of Washington with 22 different predicates, such as \texttt{advisedby/2}, \texttt{yearsinprogram/2} and  \texttt{taughtby/3}.
The aim is to predict the predicate  \texttt{advisedby/2}, namely the fact that a person is advised by another person. 

\noindent The language bias contains \textit{modeh} declarations for two-head clauses such as

\begin{small}
\begin{verbatim}
modeh(*,advisedby(+person,+person)).
\end{verbatim}
\end{small}
and \textit{modeh} declarations for multi-head clauses such as
\begin{small}
\begin{verbatim}								   		  
modeh(*,[advisedby(+person,+person),tempadvisedby(+person,+person)],
  [advisedby(A,B),tempadvisedby(A,B)],
  [professor/1,student/1,hasposition/2,inphase/2,publication/2,
  taughtby/3,ta/3,courselevel/2,yearsinprogram/2]).

modeh(*,[student(+person),professor(+person)],
  [student(P),professor(P)],
  [hasposition/2,inphase/2,taughtby/3,ta/3,courselevel/2,
  yearsinprogram/2,advisedby/2,tempadvisedby/2]).	

modeh(*,[inphase(+person,pre_quals),inphase(+person,post_quals),
  inphase(+person,post_generals)],
  [inphase(P,pre_quals),inphase(P,post_quals),inphase(P,post_generals)],
  [professor/1,student/1,taughtby/3,ta/3,courselevel/2,
  yearsinprogram/2,advisedby/2,tempadvisedby/2,hasposition/2]).	
\end{verbatim}
\end{small}
Moreover, the bias contains \textit{modeb} declarations such as 
\begin{small}
\begin{verbatim}
modeb(*,courselevel(+course, -level)).
modeb(*,courselevel(+course, #level)).
\end{verbatim}
\end{small}
An example of a two-head bottom clause that is generated from the first \textit{modeh} declaration  and the example \verb|advisedby(person155,person101)|	is

\begin{small}
\begin{verbatim}
advisedby(A,B):0.5 :- professor(B),student(A),hasposition(B,C),
  hasposition(B,faculty),inphase(A,D),inphase(A,pre_quals),
  yearsinprogram(A,E),taughtby(F,B,G),taughtby(F,B,H),taughtby(I,B,J),
  taughtby(I,B,J),taughtby(F,B,G),taughtby(F,B,H),
  ta(I,K,L),ta(F,M,H),ta(F,M,H),ta(I,K,L),ta(N,K,O),ta(N,A,P),
  ta(Q,A,P),ta(R,A,L),ta(S,A,T),ta(U,A,O),ta(U,A,O),ta(S,A,T),
  ta(R,A,L),ta(Q,A,P),ta(N,K,O),ta(N,A,P),ta(I,K,L),ta(F,M,H).
\end{verbatim}
\end{small}
An example of a multi-head bottom clause generated from the second \textit{modeh} declaration and the examples \verb|student(person218),professor(person218)| is

\begin{small}
\begin{verbatim}
student(A):0.33; professor(A):0.33 :- inphase(A,B),inphase(A,post_generals),
                                      yearsinprogram(A,C).
\end{verbatim}
\end{small}

\vspace{0.3cm}
\noindent When searching the \textit{space of clauses} for the \verb|advisedby/2| predicate, an example of a refinement from the first bottom clause  is
\begin{verbatim}
advisedby(A,B):0.5 :- professor(B).
\end{verbatim}
\emalgorithm\ is then applied to the theory composed of this single clause, using the positive and negative facts for \verb|advisedby/2| as queries for which to build the BDDs. The only parameter is updated obtaining:

\begin{verbatim}
advisedby(A,B):0.108939 :- professor(B).
\end{verbatim}
The clause is further refined to								
\begin{verbatim}
advisedby(A,B):0.108939 :- professor(B),	hasposition(B,C).	
\end{verbatim}
\vspace{0.2cm}
An example of a refinement that is generated from the second bottom clause is
\begin{verbatim}
student(A):0.33; professor(A):0.33 :- inphase(A,B).
\end{verbatim}
The updated refinement after \emalgorithm\ is
\begin{verbatim}
student(A):0.5869;professor(A):0.09832 :- inphase(A,B).
\end{verbatim}
\vspace{0.3cm}
When searching the \textit{space of theories} for the target predicate \verb|advisedby|, SLIPCOVER generates the program:
\begin{verbatim}
advisedby(A,B):0.1198 :- professor(B),inphase(A,C).
advisedby(A,B):0.1198 :- professor(B),student(A).
\end{verbatim}															
with a LL of -350.01. After EMBLEM we get:
\begin{verbatim}
advisedby(A,B):0.05465 :- professor(B),inphase(A,C).					
advisedby(A,B):0.06893 :- professor(B),student(A).
\end{verbatim}
with a LL of -318.17. Since the LL decreased, the last clause is retained and at the next iteration a new clause is added:
\begin{verbatim}
advisedby(A,B):0.12032 :- hasposition(B,C),inphase(A,D).
advisedby(A,B):0.05465 :- professor(B),inphase(A,C).					
advisedby(A,B):0.06893 :- professor(B),student(A).
\end{verbatim}

\section{Related Work}
\label{related}

Our work makes extensive use of well-known  ILP techniques: the Inverse Entailment algorithm \cite{DBLP:journals/ngc/Muggleton95} for finding the most specific clauses allowed by the language bias and the strategy for the identification of good candidate clauses. Thus SLIPCOVER is closely related to the ILP systems Progol \cite{DBLP:journals/ngc/Muggleton95} and Aleph \cite{aleph} that perform structure learning of a  logical theory by building a set of clauses iteratively.  We compare \emsalgorithm{} with Aleph in Section \ref{experiments}.

RIB \cite{RigDiM12-ML-IJ} learns the parameters of LPADs by using the information bottleneck method, an EM-like algorithm that 
was found to avoid some of the local maxima in which EM can get trapped. RIB has a good performance when the different mega-examples share the same Herbrand base. If this condition is not met, EMBLEM performs better \cite{BelRig12-IA-IJ}.

\emsalgorithm{} is an ``evolution'' of SLIPCASE  \cite{BelRig12-ILP11-IC} in terms of search strategy. SLIPCASE is based on a  simple search strategy that refines LPAD theories by  trying all possible theory revisions. SLIPCOVER instead uses bottom clauses to guide the refinement process, thus reducing the number of revisions and exploring more effectively the search space. Moreover, SLIPCOVER separates the search for promising clauses from that of the theory. By means of these modifications we have been able to get better final theories in terms of LL with respect to SLIPCASE, as shown in Section \ref{experiments}. In the following we highlight in detail the differences of the two algorithms.

SLIPCASE performs a beam search in the space of theories, starting from a trivial LPAD and using the LL of the data as the guiding heuristics. The starting theory for the beam search is user-defined: a good starting point is a theory composed of one probabilistic clause with empty body of the form \textit{$target\_predicate(\overline{V}): 0.5.$} for each target predicate, where $\overline{V}$ is a tuple of variables.
At each step of the search the theory with the highest LL is removed from the beam and a set of refinements is generated and evaluated by means of LL; then they are inserted in order of decreasing LL in the beam. The refinements of the selected theory are constructed according to a language bias based on \textit{modeh} and \textit{modeb} declarations in Progol style.  Following \cite{DBLP:journals/ai/OurstonM94,Richards95automatedrefinement} the admitted refinements are: adding or removing a literal from a clause, adding a clause with an empty body or removing a clause. 
Beam search ends when one of the following occurs: the maximum number of steps is reached, the beam is empty,  the difference between the LL of the current theory and the best previous LL drops below a threshold $\epsilon$.

SLIPCOVER  search strategy differs since it is composed of two phases:  (1) beam search in the space of clauses in order to find a set of promising clauses and (2) greedy search in the space of theories. The beam searches performed by the two algorithms differ because SLIPCOVER generates refinements of a \textit{single} clause at a time,  which are evaluated through LL (see lines 10-13 in Algorithm \ref{SEM}).
The search in the space of theories in SLIPCOVER starts from an empty theory which is iteratively extended  with one clause at a time  from those generated in the previous beam search. 
		Moreover here background clauses, the ones with a non-target predicate in the head,  are treated separately, by adding them en bloc to the best theory for target predicates.
		A further  parameter optimization step is executed and clauses that are never involved in a target predicate goal derivation are removed.
SLIPCOVER search strategy allows a more effective exploration of the search space,   resulting both in time savings and in a higher quality of the final theories, as shown by the experiments in Section \ref{experiments}.

Previous works on learning the structure of probabilistic logic programs include 
\cite{DBLP:conf/ilp/KerstingR08}, that proposed a scheme for learning both the probabilities and the structure of Bayesian logic programs by combining techniques from the learning from interpretations setting of ILP with score-based techniques for learning Bayesian networks. We share with this approach the scoring function, the LL of the data given a candidate structure and the greedy search in the space of structures.

 \citeANP{DBLP:journals/ml/RaedtKKRT08}  \citeyear{DBLP:journals/ml/RaedtKKRT08} presented an algorithm for performing theory compression on ProbLog programs.
Theory compression means removing as many clauses as possible from the theory in order to maximize the likelihood w.r.t. a set of positive and negative examples. No new clause can be added to the theory. 

\citeANP{DBLP:conf/ilp/RaedtT10} \citeyear{DBLP:conf/ilp/RaedtT10} introduced the probabilistic rule learner ProbFOIL, which combines the rule learner FOIL \cite{DBLP:conf/ecml/QuinlanC93} with ProbLog  \cite{DBLP:conf/ijcai/RaedtKT07}. Logical rules are learned from probabilistic data in the sense that both the examples themselves and their classifications can be probabilistic. The set of rules has to allow to predict the probability of the examples from their description. In this setting  the parameters
(the probability values) are fixed and the structure (the rules)
are to be learned.

LLPAD \cite{Rig04-ILP04-IC} and ALLPAD \cite{Rig-ILP06-IC,DBLP:journals/ml/Riguzzi08} learn ground LPADs by first generating a set of candidate clauses satisfying certain constraints and then
solving an integer linear programming model to select a subset of the clauses that assigns the given probabilities to the examples. While LLPAD looks for a perfect match, ALLPAD looks for a solution that minimizes the difference of the learned and given probabilities of the examples. In both cases  the learned clauses are restricted to have mutually exclusive bodies.

SEM-CP-logic \cite{DBLP:journals/fuin/MeertSB08} learns parameters and  structure of ground CP-logic programs. It performs learning by considering the Bayesian networks equivalent to  CP-logic programs and by applying techniques for learning Bayesian networks. In particular, it applies the Structural Expectation Maximization (SEM) algorithm \cite{DBLP:conf/uai/Friedman98}: it  iteratively generates refinements of the equivalent Bayesian network  and it  greedily chooses the one that maximizes the BIC score \cite{BIC}.
In SLIPCOVER, we used the LL as a score because experiments with BIC were giving inferior results. Moreover, SLIPCOVER differs from SEM-CP-logic also because  it searches the clause space and it refines clauses with standard ILP refinement operators, which allow to learn non ground theories.

\citeANP{Getoor+al:SRL07} \citeyear{Getoor+al:SRL07} described a comprehensive framework for learning statistical models called Probabilistic Relational Models (PRMs).  These extend Bayesian networks with the concepts of objects, their properties, and relations between them, and specify a template for a probability distribution over a database. The template includes a relational component, that describes the relational schema for the domain, and a probabilistic component, that describes the probabilistic dependencies that hold in it. A method for the automatic construction of a PRM from an existing database is shown, together with parameter estimation, structure scoring criteria  and a definition of the model search space.

\citeANP{DBLP:conf/uai/CostaPQC03} \citeyear{DBLP:conf/uai/CostaPQC03} presented an
extension of logic programs that makes it possible
to specify a joint probability distribution
over missing values in a database or logic program, in analogy to PRMs. This extension is based on constraint logic programming (CLP) and is called CLP(BN). Existing ILP
systems like Aleph can be used to learn CLP(BN) programs with simple modifications. 

\citeANP{PaesRZC06} \citeyear{PaesRZC06} described the first theory revision system for SRL, \textit{PFORTE} for ``Probabilistic First-Order
Revision of Theories from Examples'', which starts from an approximate initial theory and applies modifications in places that performed badly in classification.  PFORTE uses a two-step approach. The completeness component uses generalization operators to address failed proofs and the classification component addresses classification problems using generalization and specialization operators. It is presented as an alternative  to  algorithms that learn from scratch.

Structure learning has been thoroughly investigated for Markov Logic: in \cite{DBLP:conf/icml/KokD05} the authors proposed two approaches. The first is a beam search that adds a clause at a time to the theory using 
weighted pseudo-likelihood as a scoring function. The second is called  shortest-first search and adds the $k$ best  clauses of length $l$ before considering clauses with length $l+1$.

\citeANP{DBLP:conf/icml/MihalkovaM07} \citeyear{DBLP:conf/icml/MihalkovaM07}  proposed a bottom-up algorithm for learning Markov Logic Networks, called BUSL, that is based on relational pathfinding: paths of true ground atoms that are linked via
their arguments are found and generalized into first-order rules. 

\citeANP{DBLP:conf/icml/HuynhM08} \citeyear{DBLP:conf/icml/HuynhM08} introduced a two-step method for inducing the structure of MLNs: (1) learning a large number of promising clauses through a specific configuration of Aleph (ALEPH++), followed by (2) the application of a new discriminative MLN parameter learning algorithm. This algorithm differs from the standard weight learning one \cite{Lowd07efficientweight} in the use of an exact probabilistic inference method and of a L1-regularization of the parameters, in order to encourage assigning low weights to clauses. The complete method is called ALEPH++ExactL1; we compare \emsalgorithm{} with it in Section \ref{experiments}.

In \cite{DBLP:conf/icml/KokD09}, the structure of Markov Logic theories is learned by applying a generalization of relational pathfinding. A database is viewed as a hypergraph with constants as nodes and true ground atoms
as hyperedges. Each hyperedge is labeled with a predicate symbol.
First a hypergraph over clusters of constants is found, then pathfinding is applied on this ``lifted'' hypergraph. The resulting algorithm is called LHL.

 \citeANP{DBLP:conf/icml/KokD10}  \citeyear{DBLP:conf/icml/KokD10} presented the  algorithm ``Learning Markov Logic Networks using  Structural Motifs'' (LSM). It is based on the observation that 
relational data frequently contain recurring patterns
of densely connected objects called \emph{structural motifs}. LSM limits the 
search to these patterns. Like LHL, LSM views a database as a hypergraph and groups nodes that are densely connected by many paths and the hyperedges connecting
the nodes into a motif. Then it evaluates whether the motif appears frequently enough in the data and finally it applies relational pathfinding to find rules. This process, called \emph{createrules} step, is followed by weight learning with the Alchemy system. LSM was experimented on various datasets and found to be superior to other methods, thus representing the state of the art in Markov Logic Networks' structure learning and in SRL in general. We compare \emsalgorithm{} with LSM in Section \ref{experiments}.

A different approach is taken in \cite{DBLP:conf/ilp/BibaFE08} where the algorithm DSL is presented, that performs discriminative structure learning by repeatedly adding a clause to the theory through iterated local search, which performs a walk in the space of local optima.
We share with this approach the discriminative nature of the algorithm and the scoring function.

\section{Experiments}
\label{experiments}
SLIPCOVER has been tested on five real world datasets: HIV,  UW\--CSE, WebKB, Mutagenesis and Hepatitis.

\subsection{Datasets}

\paragraph{HIV}
The HIV dataset\footnote{Kindly provided by Wannes Meert.} \cite{DBLP:journals/jcb/BeerenwinkelRDHKSL05}  records mutations in HIV's reverse transcriptase gene in patients that are treated with the drug zidovudine. It contains 364 examples, 
each of which specifies the presence or not of six classical zidovudine mutations, denoted by the atoms: \verb|41L|, \verb|67N|, \verb|70R|, \verb|210W|, \verb|215FY| and \verb|219EQ|. These atoms indicate the location where the mutation occurred (e.g., \textit{41}) and the amino acid to which the position mutated (e.g., \textit{L} for Leucine).
The goal is to discover causal relationships between the occurrences of mutations in the virus, so all the predicates are set as target.

\paragraph{UW-CSE}
The UW-CSE dataset\footnote{\texttt{http://alchemy.cs.washington.edu/data/uw-cse}} \cite{DBLP:conf/icml/KokD05} contains information about the Computer Science department of the University of Washington, and is split into five mega\--examples, each containing facts for a particular research area. 
The goal is to predict the \texttt{advisedby(X,Y)} predicate, namely the fact that a person \verb|X| is advised by another person \verb|Y|, so this represents the target predicate.

\paragraph{WebKB}
The WebKB dataset\footnote{\url{http://alchemy.cs.washington.edu/data/webkb}} describes web pages from the computer science
departments of four universities. We used the version of the dataset
from \cite{DBLP:journals/ml/CravenS01} that contains 4,165 web pages and 10,935 web
links, along with words on the web pages.
Each web page \verb|P| is labeled with some subset of the categories:  student, faculty, research project and course. The goal is to
predict these categories from the web pages' words and link structures.

\paragraph{Mutagenesis}
The Mutagenesis dataset\footnote{\url{http://www.doc.ic.ac.uk/~shm/mutagenesis.html}} \cite{DBLP:journals/ai/SrinivasanMSK96} contains information about a number of aromatic and heteroaromatic nitro drugs, including their chemical structures in terms of atoms, bonds and a number of molecular substructures such as five- and six-membered rings, benzenes, phenantrenes and others. The fundamental Prolog facts are \verb|bond(compound,atom1,atom2,bondtype)| - stating that in the \textit{compound} a  bond of type \textit{bondtype} can be found between the atoms \textit{atom1} and \textit{atom2} - and \verb|atm(compound,atom,element,atomtype,charge)|, stating that \textit{compound}'s \textit{atom} is of element \textit{element}, is of type \textit{atomtype}  and has partial charge \textit{charge}. From these facts many elementary molecular substructures can be defined, and we used the tabulation of these, available in the dataset, rather than the clause definitions based on \texttt{bond/4} and \texttt{atm/5}. This greatly sped up learning.

The problem here is to predict the mutagenicity of the drugs. The prediction of mutagenesis is important as it is relevant to the understanding and prediction of carcinogenesis. 
The subset of the compounds having positive levels of log mutagenicity are labeled ``active'' and constitute the positive examples, the remaining ones are  ``inactive'' and constitute the negative examples. 

The data is split into two subsets (188+42 examples). We considered  the first one, composed of 125 positive and 63 negative compounds.
The goal is to predict if a drug is active, so the target predicate is \texttt{active(drug)}.

\paragraph{Hepatitis}
The Hepatitis dataset\footnote{\url{http://www.cs.sfu.ca/~oschulte/jbn/dataset.html}} \cite{DBLP:journals/ml/KhosraviSHG12}  is derived from the PKDD02 Discovery
Challenge database \cite{PKDD02-DiscChal}. It contains information on the laboratory examinations of hepatitis B and C infected patients. Seven tables are used to store this information. The goal is to predict the type of hepatitis of a patient, so the target predicate is \texttt{type(pat,type)} where \texttt{type} can be \texttt{type\_b} or \texttt{type\_c}. We generated negative examples for \texttt{type/2} by adding, for each fact \texttt{type(pat,type\_b)}, the fact \texttt{neg(type(pat,type\_c))}  and for each fact \texttt{type(pat,type\_c)}, the fact \texttt{neg(type(pat,type\_b))}.

\vspace{0.4cm}
Statistics on all the domains are reported in Table \ref{tab:charact}. The number of negative testing examples is sometimes different from that of negative training examples because, while in training we explicitly provide negative examples, in testing we consider all the ground instantiations of the target predicates that are not positive as negative.

\begin{table}[ht]
    \caption{Characteristics of the datasets for the experiments: target predicates, number of constants, of predicates, of tuples (ground atoms), of positive and negative training and testing examples for target predicate(s), of folds. The number of tuples includes the target positive examples.} 
    \label{tab:charact}
    \centering
{\footnotesize
\begin{tabular}%
{lp{2.3cm}p{0.6cm}p{0.5cm}p{0.7cm}p{0.8cm}p{0.9cm}p{0.8cm}p{0.7cm}} \hline 
Dataset&Target 
Pred&Const&Preds&Tuples&Pos.Ex.&Train.&Test.&Folds\\ 
    &&&&&&Neg. Ex.&Neg. Ex.&\\ \hline
HIV  &41L,67N,70R, &0	          &6			  	&2184				 &590				&1594	      &1594						&5\\ 
  &210W,215FY,\\
  &219EQ \\ \hline
UW-CSE &advisedby(X,Y)  													 &1323 				&15 				&2673				 &113				&4079          &16601					&5 \\ \hline
WebKB &coursePage(P) &4942 &8 &290973 &1039 &15629   &16249   &4 \\ 
&facultyPage(P)\\ 
&researchPrPage(P)\\
&studentPage(P)\\ \hline
Mutagenesis &active(D) 														 &7045 				&20 				&15249		   &125					       &63    &63			      	&10 \\ \hline
Hepatitis &type(X,T) 															 &6491 			  &19 				&71597 			 &500			  	       &500      &500						&5  \\ \hline
\end{tabular}
}
\end{table}

\subsection{Methodology}
\emsalgorithm\ is implemented  in Yap Prolog \cite{Yap} and is compared  with Aleph, SLIPCASE and SEM-CP-logic for probabilistic logic programs, and  LSM and ALEPH++ExactL1 for Markov Logic Networks.\\
All experiments were performed on Linux machines with an Intel Core 2 Duo E6550 (2333 MHz) processor and 4 GB of RAM.

\subsubsection{Parameter settings}

\paragraph{SLIPCOVER and SLIPCASE}\mbox{}\\
\emsalgorithm{} offers the following parameters: the number {\small$NInt$} of mega-examples on which to build the bottom clauses, the number $NA$ of bottom clauses to be built for each mega-example, the number $NS$ of saturation steps, the maximum number $NI$ of clause search iterations, the size $NB$ of the beam, the maximum number $NV$ of variables in a rule, 
the maximum numbers $NTC$ and $NBC$  of target and background clauses respectively, the semantics (standard or approximate) and 
the additional parameters $D$, $NEM$, $\epsilon$ and $\delta$ for \textsc{EMBLEM}.

SLIPCASE offers the following parameters: $NIT$, the number of theory revision iterations,  $NR$, the maximum number of rules in a learned theory, $\epsilon_s$ and $\delta_s$, respectively the minimum difference and relative difference between the LL of the theory in two refinement iterations, and finally EMBLEM's parameters.
The parameters  $NV$, $NB$ and the semantics are shared with \emsalgorithm{}.

\vspace{0.3cm}
For \emalgorithm{}  we set $\epsilon=10^{-4}$, $\delta=10^{-5}$ and $NEM=+\infty$, since we observed that it usually converged quickly. 

For SLIPCASE we set $\epsilon_s =10^{-4}$ and $\delta_s=10^{-5}$ in all experiments except Mutagenesis, where we used $\epsilon_s =10^{-20}$ and $\delta_s=10^{-20}$.

For SLIPCOVER we always set $NS=1$  to limit the size of the bottom clauses.%

All the other parameters of SLIPCOVER and SLIPCASE have been chosen to avoid lack of memory errors and to keep computation time within 24 hours. This is true also for the depth bound $D$ used in domains where the language bias allowed recursive clauses. The values we used for $D$ are 2 or 3; when the theory in not cyclic this parameter is not relevant. 

The parameter settings for SLIPCOVER and SLIPCASE on the domains  can be found in Tables  \ref{tab:parSO} and \ref{parSC} respectively.

\begin{table}[ht]
    \caption{Parameter setting for the experiments with SLIPCOVER. `-' means the parameter is not relevant.} 
    \label{tab:parSO}
    \centering
{\small
\begin{tabular}%
{lcccccccccc} \hline 
Dataset 	&NInt &NS &NA &NI &NV   &NB  &NTC & NBC 	&D  &semantics \\ \hline
HIV  			&1  	&1 	&1	&10	& -		&10	 &50 & -		  &3	&approximate\\ \hline   %
UW-CSE    &4	  &1 &1 &10  &4   &100 &10000 &50    &3  &approximate\\ \hline
WebKB 		&1	  &1&1  &5   &4   &15  &50 & -       &2  &standard \\ \hline
Mutagenesis &1	&1  &1&10  &5   &20	 &100 & -      &-  &standard\\ \hline
Hepatitis &1		&1  &1&10  &5   &20	 &1000 & -			&- 	&approximate\\ \hline
\end{tabular}
}
\end{table}

\begin{table}[ht]
    \caption{Parameter setting for the experiments with SLIPCASE. `-' means the parameter is not relevant.} 
    \label{parSC}
    \centering
{\small
\begin{tabular}%
{lcccccc} \hline 
Dataset & NIT & NV &NB &NR  &D   &semantics\\ \hline
HIV 		 &10	&- 	&5 	&10		&3   &standard \\ \hline
UW-CSE   &10	&5	&20	&10		&- 	&standard \\ \hline
WebKB 	 &10 	&5	&20	&10		&-  &approximate\\ \hline
Mutagenesis&10&5  &20 &10		&-  &standard\\ \hline
Hepatitis&10  &5  &20 &10   &-  &standard\\ \hline
\end{tabular}
}
\end{table}

\paragraph{Aleph} \mbox{}\\
We modified the standard settings as follows: the maximum number of literals in a clause was set to 7 (instead of the default 4) for UW-CSE and Mutagenesis, since here clause bodies are generally long. The minimum number of positive examples covered by an acceptable clause was set to 2, as suggested by the system manual \cite{aleph}.  The search strategy was forced to continue until all remaining elements in the search space were definitely worse than the current best element (normally, search would stop when all remaining elements are no better than the current best), by setting the \texttt{explore} parameter to \verb|true|. The \verb|induce| command was used to learn the clauses.

We report results only for UW-CSE, WebKb and Mutagenesis, since on HIV and Hepatitis Aleph returned the set of examples as the final theory, not being able to find good enough generalizations.

\paragraph{SEM-CP-logic}\mbox{}\\
We report the results only on HIV as the system learns only ground theories and the other datasets require theories with variables.

\paragraph{LSM}\mbox{}\\ The weight learning step can be generative or discriminative, according to whether  the aim is to accurately predict all or a specific predicate respectively; for the discriminative case we used the preconditioned scaled conjugate gradient technique, because it was found to be the state of the art \cite{Lowd07efficientweight}.

\paragraph{ALEPH++ExactL1} \mbox{}\\
We used the \verb|induce_cover| command and the parameter settings for Aleph specified in \cite{DBLP:conf/icml/HuynhM08} on the datasets on which Aleph could return a theory.

\subsubsection{Test}

For testing on HIV, we used a five-fold cross-validation. We computed the probability of each mutation in each example given the value of the remaining mutations. The presence of a mutation in an example is considered as a positive example, while its absence as a negative example.

For testing on UW-CSE and Hepatitis we applied a five-fold cross-validation; on
WebKB  a four-fold cross-validation, on Mutagenesis  a ten-fold cross-validation.

We drew a Precision-Recall curve and a Receiver Operating Characteristics curve and computed the Area Under the Curve (AUCPR
and AUCROC respectively) using the methods reported in \cite{DBLP:conf/icml/DavisG06,DBLP:journals/prl/Fawcett06}. 
Recently, \citeANP{DBLP:conf/icml/BoydDPC12} \citeyear{DBLP:conf/icml/BoydDPC12} showed that, when the skew is larger than 0.5, the AUCPR is not adequate to evaluate the performance of learning algorithms, where the skew is the ratio between the number of positive examples and the total number of examples. Since for Mutagenesis and Hepatitis the skew is close to 0.5, for these datasets we computed the Normalized Area Under the PR Curve (AUCNPR) proposed in \cite{DBLP:conf/icml/BoydDPC12}.

In the case of  Aleph tests, we annotated the head of each learned clause with probability 0.5 before testing 
in order to turn the sharp logical classifier into a probabilistic one and to assign higher probability to those examples that have more successful derivations.

Tables \ref{tab:pr} and \ref{tab:roc} show respectively the AUCPR and AUCROC averaged over the folds for all algorithms and datasets. Table \ref{tab:npr} shows the AUCNPR for all algorithms on the  Mutagenesis and Hepatitis datasets, while Table \ref{tab:times} shows the learning times in hours. 

Tables \ref{tab:p-valuePR} and \ref{tab:p-valueROC} show the p-value of a paired two-tailed t-test at the 5\% significance level  of the difference in AUCPR and AUCROC between \emsalgorithm{} and  SLIPCASE/SEM-CP-logic/Aleph/LSM/ALEPH++ExactL1 on all datasets (significant differences in favor of \emsalgorithm{} in bold).

\begin{figure}[t]
\centering
	\includegraphics[width=.6\textwidth]{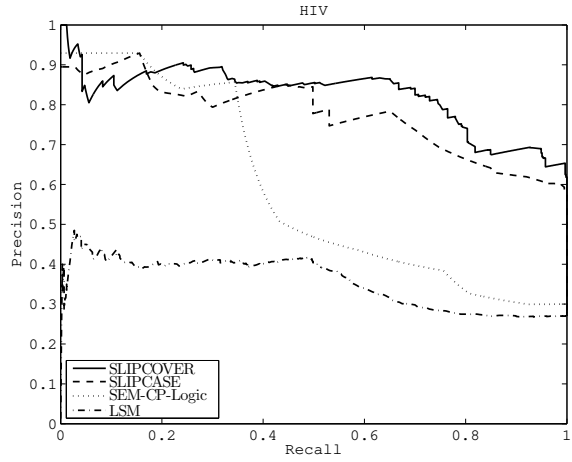}
\caption{\label{hiv_pr}PR  curves for HIV.}
\end{figure}

\begin{figure}[t]
\centering
\includegraphics[width=.6\textwidth]{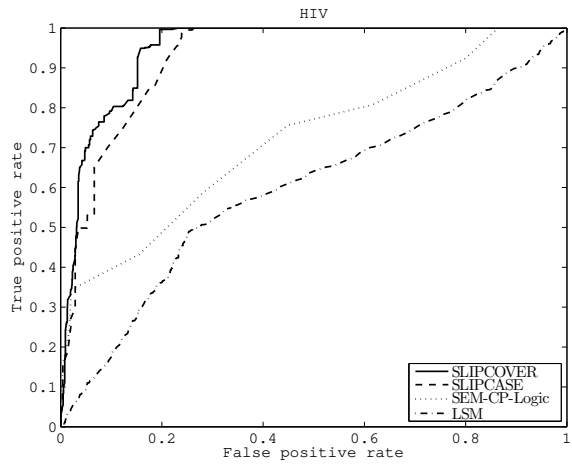}
\caption{\label{hiv_roc}ROC curves for HIV.}
\end{figure}

\begin{figure}[t]
\centering
	\includegraphics[width=.6\textwidth]{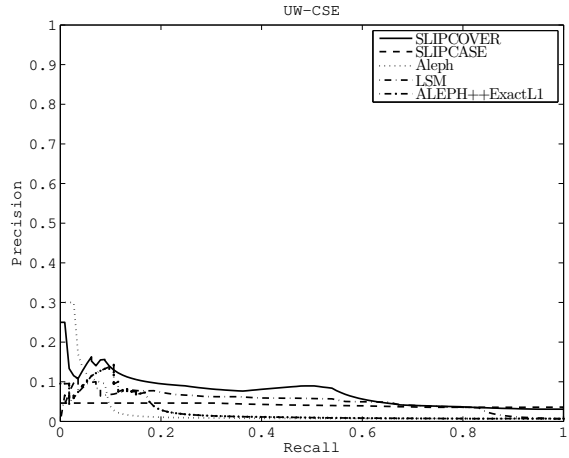}
\caption{\label{uwcse_pr}PR  curves for UWCSE.}
\end{figure}

\begin{figure}[t]
\centering
\includegraphics[width=.6\textwidth]{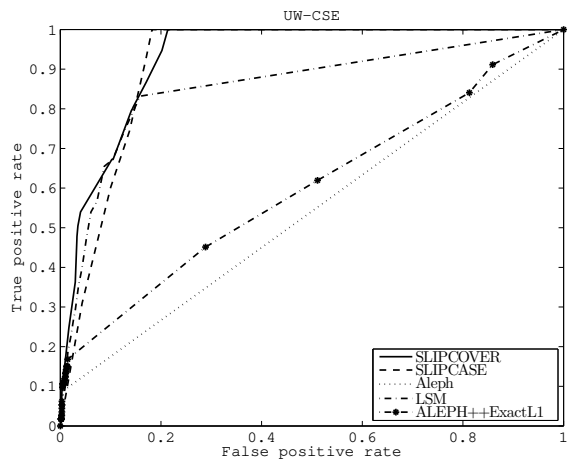}
\caption{\label{uwcse_roc}ROC curves for UWCSE.}
\end{figure}

\begin{figure}[t]
\centering
	\includegraphics[width=.6\textwidth]{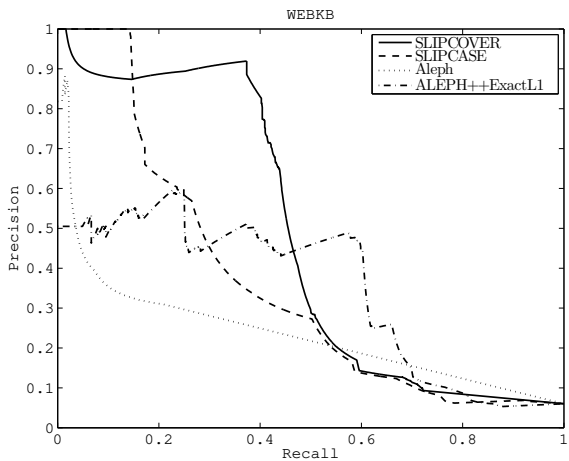}
\caption{\label{webkb_pr}PR  curves for WebKB.}
\end{figure}

\begin{figure}[t]
\centering
\includegraphics[width=.6\textwidth]{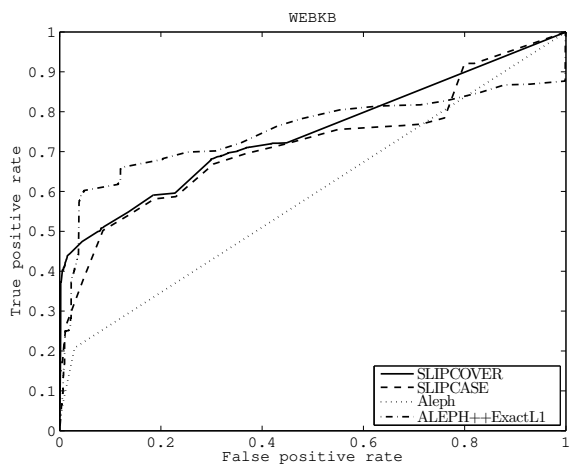}
\caption{\label{webkb_roc}ROC curves for WebKB.}
\end{figure}

\begin{figure}[t]
\centering
	\includegraphics[width=.6\textwidth]{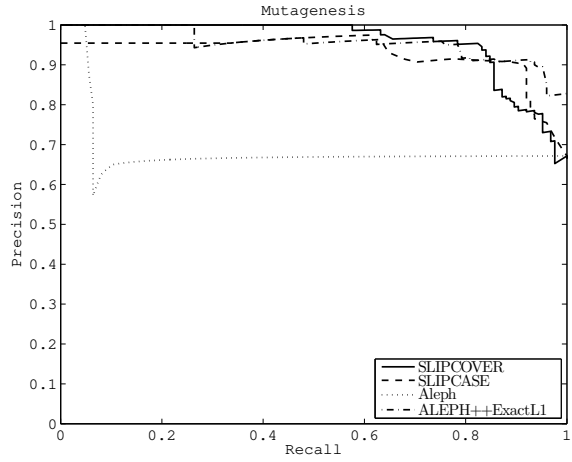}
\caption{\label{mutagenesis_pr}PR  curves for Mutagenesis.}
\end{figure}

\begin{figure}[t]
\centering
\includegraphics[width=.6\textwidth]{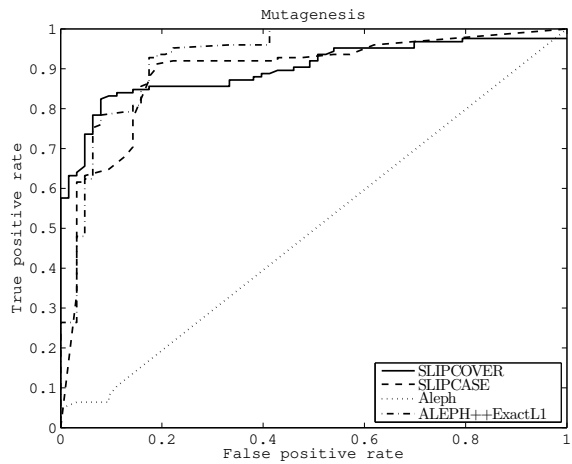}
\caption{\label{mutagenesis_roc}ROC curves for Mutagenesis.}
\end{figure}

\begin{figure}[t]
\centering
	\includegraphics[width=.6\textwidth]{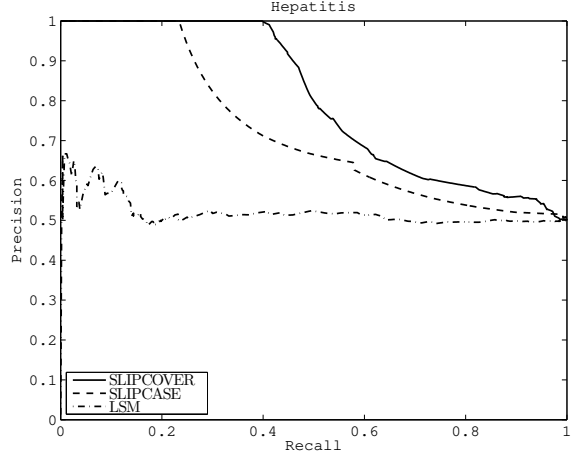}
\caption{\label{hepatitis_pr}PR  curves for Hepatitis.}
\end{figure}

\begin{figure}[t]
\centering
\includegraphics[width=.6\textwidth]{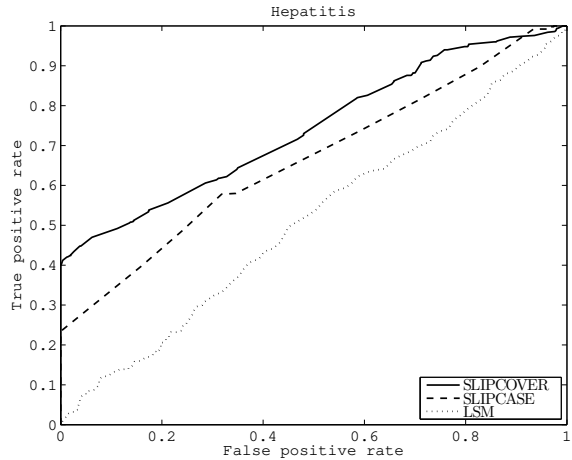}
\caption{\label{hepatitis_roc}ROC curves for Hepatitis.}
\end{figure}

Figures \ref{hiv_pr}, \ref{uwcse_pr}, \ref{webkb_pr}, \ref{mutagenesis_pr} and \ref{hepatitis_pr} show the PR curves for all datasets, while Figures \ref{hiv_roc}, \ref{uwcse_roc}, \ref{webkb_roc}, \ref{mutagenesis_roc} and \ref{hepatitis_roc} show ROC curves. These curves have been obtained by collecting the testing examples, together with the probabilities assigned to them in testing, in a single set and then building the curves with the methods of \cite{DBLP:conf/icml/DavisG06,DBLP:journals/prl/Fawcett06}.

\subsubsection{Results}

In all the datasets negative literals are not allowed in clauses' bodies, so the LPADs learnt are surely sound.

\paragraph{HIV}
The language bias for SLIPCOVER and SLIPCASE allowed each atom to appear in the head and in the body (cyclic theory).
For SLIPCOVER, $NBC$ was not relevant since all predicates are target. 
The input  theory for SLIPCASE was composed of a probabilistic clause of the form \texttt{<mutation>:0.2} for each of the six mutations. 
The  clauses of the final theories are characterized by one head atom (SLIPCOVER) or one/two head atoms (SLIPCASE).

For SEM-CP-logic, we tested the learned theory  reported in \cite{DBLP:journals/fuin/MeertSB08} over each of the five folds.%

For LSM, we used the generative training algorithm to learn weights, because all the predicates are target, and  the MC-SAT algorithm for inference over the test fold, by specifying all the six mutations as query atoms. 
\vspace{0.2cm}

On this dataset SLIPCOVER achieves  higher areas with respect to SLIPCASE, SEM-CP-logic and LSM.


The medical literature states that \verb|41L|, \verb|215FY| and \verb|210W| tend to occur together, and that \verb|70R| and \verb|219EQ| tend to occur together as well. SLIPCASE and LSM find only one of these two connections and the simple MLN learned by LSM may explain its low AUCs.
SLIPCOVER instead learns many more clauses where both connections are found, with probabilities larger than the other clauses. The longer learning time with respect to the other systems mainly depends on the theory search phase, since the $TC$ list can contain up to 50 clauses and the final theories have on average 40, so many theory refinement steps are executed.

In the following we show examples of rules that have been learned by the systems, showing only rules expressing the above connections. 

\noindent SLIPCOVER learned the clauses:

\begin{small}
\begin{verbatim}
70R:0.950175 :- 219EQ.    									
41L:0.24228	 :- 215FY,	210W.
41L:0.660481 :- 210W.
41L:0.579041 :- 215FY.
219EQ:0.470453 :- 67N,	70R.
219EQ:0.400532 :- 70R.
215FY:0.795429 :- 210W,	219EQ.
215FY:0.486133 :- 41L,	219EQ.
215FY:0.738664 :- 67N,	210W.
215FY:0.492516 :- 67N,	41L.
215FY:0.475875 :- 210W.
215FY:0.924251 :- 41L.
210W:0.425764  :- 41L.
\end{verbatim}
\end{small}
SLIPCASE instead learned:
\begin{small}
\begin{verbatim}
41L:0.68 :- 215FY.
215FY:0.95 ; 41L:0.05 :- 41L.
210W:0.38 ; 41L:0.25  :- 41L, 215FY.
\end{verbatim}
\end{small}
The clauses learned by SEM-CP-Logic that include the two connections are:
\begin{small}
\begin{verbatim}
70R:0.30   :-  219EQ.
215FY:0.90 :-  41L.
210W:0.01  :-  215FY.
\end{verbatim}
\end{small}
LSM learned:
\begin{small}
\begin{verbatim}
1.19 !g41L(a1) v g215FY(a1)
0.28  g41L(a1) v !g215FY(a1)
\end{verbatim}
\end{small}

\paragraph{UW-CSE}
The language bias for SLIPCOVER allowed all predicates to appear in the head and in the body of clauses; all except \texttt{advisedby/2} are background predicates. Moreover, nine \textit{modeh} facts declare disjunctive heads, three of them are shown in Section \ref{slipcover}. These \textit{modeh} declarations have been defined by looking at the hand crafted theory used for parameter learning in \cite{BelRig13-IDA-IJ}: for each disjunctive clause in the theory, a \textit{modeh} fact is derived. On this dataset head refinement is applied. The approximate semantics has been used to limit learning time.

The language bias for SLIPCASE allowed \verb|advisedby/2| to appear in the head only  and all the other predicates in the body only; for this reason we ran it with no depth bound.
The input theory was composed of two clauses of the form \texttt{advisedby(X,Y):0.5.}

For LSM, we used the discriminative  training  algorithm for learning the weights, by  specifying \texttt{advisedby/2} as the only non\--evidence predicate, and
 the MC-SAT algorithm for inference over the test folds, by  specifying \texttt{advisedby/2} as the query predicate.

For Aleph and ALEPH++ExactL1 the same language bias as SLIPCASE was used.
\vspace{0.2cm}

On this dataset SLIPCOVER achieves  higher AUCPR  and  AUCROC than all other systems. This is a difficult dataset, as testified by the low values of areas achieved by all systems, and represents a challenge for structure learning algorithms.


SLIPCASE learns simple programs composed of a single clause per fold; this also explains the low learning time. In two folds out of five it learns the theory

\begin{small}
\begin{verbatim}
advisedby(A,B):0.26 :- professor(B), student(A).
\end{verbatim}
\end{small}
An example of a theory learned by LSM is
\begin{small}
\begin{verbatim} 
3.77122  professor(a1) v !advisedBy(a2,a1)
0.03506  !professor(a1) v !advisedBy(a2,a1)
2.27866  student(a1) v !advisedBy(a1,a2)
1.25204  !student(a1) v !advisedBy(a1,a2)
0.64834  hasPosition(a1,a2) v !advisedBy(a3,a1)
1.23174  !advisedBy(a1,a2) v inPhase(a1,a3)
\end{verbatim}
\end{small}
SLIPCOVER learns theories able to better model the domain, at the expense of a longer learning time than SLIPCASE.
Examples of clauses are:

\begin{small}
\begin{verbatim}
advisedby(A,B):0.538049 ; tempadvisedby(A,B):0.261159 :- ta(C,A,D),
                                                         taughtby(C,B,D).
advisedby(A,B):2.97361e-10 :- publication(C,A).
advisedby(A,B):0.0396684 :- publication(C,B),publication(D,A),
                            professor(B),student(A).
advisedby(A,B):0.0223801 :- publication(C,A),publication(D,A),
                            professor(B),student(A).
advisedby(A,B):0.052342 :- professor(B).
hasposition(A,faculty):0.344719; hasposition(A,faculty_adjunct):0.225888; 
  hasposition(A,faculty_emeritus):0.14802; 
  hasposition(A,faculty_visiting):0.0969946 :- professor(A).
professor(A):0.569775 :- hasposition(A,B).
...
\end{verbatim}    
\end{small}
Aleph and ALEPH++ExactL1 mainly differ in the number of learned clauses, while body literals and their ground arguments are essentially the same. \\
ALEPH++ExactL1 returns more complex  MLNs than LSM, and performs slightly better.

\paragraph{WebKB}
The language bias for SLIPCOVER and SLIPCASE allowed predicates representing the four categories both in the head and in the body of clauses. Moreover, the body could contain the atom \verb|linkTo(_Id,Page1,Page2)|  (linking two pages) and the atom \verb|has(word,Page)| where \verb|word| is a constant representing a word appearing in the pages. \\
The input theory for SLIPCASE was composed of clauses of the form\\
\verb|<category>Page(Page):0.5.|,
one for each category. The target predicates were treated as closed world in this case, so the corresponding literals in the clauses' body are resolved only with examples in the mega-examples and not with the other clauses in the theory to limit execution time, therefore $D$ is not relevant. The approximate semantics was used for SLIPCASE to limit the learning time.

LSM failed on this dataset because the weight learning phase quickly exhausted the available memory on our machines (4 GB). This dataset is in fact quite large, with 15 MB input files on average.

For Aleph and ALEPH++ExactL1, we overcame the limit of one target predicate per run by executing Aleph four times on each fold, once for each target predicate. In each run, we removed the target predicate from the \textit{modeb} declarations to prevent Aleph from testing cyclic theories and going into a loop. %
\vspace{0.2cm}

On this dataset SLIPCOVER achieves  higher AUCPR  and AUCROC than the other systems but the differences are not statistically significant.
 

A fragment of a theory learned by SLIPCOVER is:

\begin{small}
\begin{verbatim} 
studentPage(A):0.9398:- linkTo(B,C,A),has(paul,C),has(jame,C),has(link,C).
researchProjectPage(A):0.0321475:- linkTo(B,C,A),has(project,C),
                                   has(depart,C),has(nov,A),has(research,C).
facultyPage(A):0.436275 :- has(professor,A),has(comput,A).
coursePage(A):0.0630934 :- has(date,A),has(gmt,A).
\end{verbatim}
\end{small}
Aleph and ALEPH++ExactL1 learned many more clauses for every target predicate than SLIPCOVER. For \texttt{coursePage} for example ALEPH++ExactL1 learned the clauses

\begin{small}
\begin{verbatim} 
coursePage(A) :-   has(file,A), has(instructor,A), has(mime,A).
coursePage(A) :-   linkTo(B,C,A), has(digit,C), has(theorem,C).
coursePage(A) :-   has(instructor,A), has(thu,A).
coursePage(A) :-   linkTo(B,A,C), has(sourc,C), has(syllabu,A).
coursePage(A) :-   linkTo(B,A,C), has(homework,C), has(syllabu,A).
coursePage(A) :-   has(adapt,A), has(handout,A).
coursePage(A) :-   has(examin,A), has(instructor,A), has(order,A).
coursePage(A) :-   has(instructor,A), has(vector,A).
coursePage(A) :-   linkTo(B,C,A), has(theori,C), has(syllabu,A).
coursePage(A) :-   linkTo(B,C,A), has(zpl,C), has(topic,A).
coursePage(A) :-   linkTo(B,C,A), has(theori,C), has(homework,A).
coursePage(A) :-   has(decemb,A), has(instructor,A), has(structur,A).
coursePage(A) :-   has(apr,A), has(client,A), has(cours,A).
coursePage(A) :-   has(home,A), has(spring,A), has(syllabu,A).
coursePage(A) :-   has(ad,A), has(copyright,A), has(cse,A).
   ...
\end{verbatim}
\end{small}
In this domain SLIPCASE learns fewer and simpler clauses (many  with an empty body) for each fold than SLIPCOVER.
 Moreover,	SLIPCASE search strategy generates thousands of refinements for each theory extracted from the beam, while SLIPCOVER beam search generates less than a hundred refinements from each bottom clause, thus achieving a lower learning time.

\paragraph{Mutagenesis}

The language bias for SLIPCOVER and SLIPCASE allowed \texttt{active/1} only  in the head, so the depth $D$ was not relevant.
For SLIPCASE, we set $\epsilon_s =10^{-20}$ and $\delta_s=10^{-20}$ to ensure that it performed 10 refinement iterations, so that its execution time was close to that of SLIPCOVER. 
The input theory for SLIPCASE contained two facts of the form  \texttt{active(D):0.5}.

LSM failed on this dataset because the structure learning phase (\textit{createrules} step) quickly gave a memory allocation error when generating \verb|bond/4| groundings.
\vspace{0.2cm}

On this dataset SLIPCOVER achieves  higher AUCPR  and AUCROC than the other systems,  except ALEPH++ExactL1, which achieves the same AUCPR as SLIPCOVER and non statistically significant higher AUCROC. The differences between SLIPCOVER and Aleph are instead statistically significant.


SLIPCOVER learns more complex programs with respect to those learned by SLIPCASE, that contain only two or three clauses for each fold. 
	
\citeANP{Muggleton94mutagenesis:ilp} \citeyear{Muggleton94mutagenesis:ilp} report the results of the application of Progol to this dataset. In the following we present the clauses learned by Progol paired with the most similar clauses learned by SLIPCOVER and ALEPH++ExactL1.

Progol learned

\begin{small}
\begin{verbatim}
active(A) :- atm(A,B,c,10,C),atm(A,D,c,22,E),bond(A,D,B,1).
\end{verbatim}
\end{small}
where a carbon atom \textit{c} of type 22 is known to be in an aromatic ring.\\
SLIPCOVER learned

\begin{small}
\begin{verbatim}
active(A):9.41508e-06 :- bond(A,B,C,7), atm(A,D,c,22,E).		        
active(A):1.14234e-05 :- benzene(A,B), atm(A,C,c,22,D).
\end{verbatim}
\end{small}
where a bond of type 7 is an aromatic bond and benzene is a 6-membered carbon aromatic ring.

Progol learned

\begin{small}
\begin{verbatim}
active(A) :- atm(A,B,o,40,C), atm(A,D,n,32,C).
\end{verbatim}
\end{small}
SLIPCOVER instead learned:
\begin{small}
\begin{verbatim}
active(A):5.3723e-04 :-  bond(A,B,C,7), atm(A,D,n,32,E).
\end{verbatim}
\end{small}

The clause learned by Progol 
\begin{small}
\begin{verbatim}
active(A):-  atm(A,B,c,27,C),bond(A,D,E,1),bond(A,B,E,7).
\end{verbatim}
\end{small}
where a carbon atom \textit{c} of type 27 merges 2 6-membered aromatic rings, is similar to SLIPCOVER's

\begin{small}
\begin{verbatim}
active(A):0.135014 :- benzene(A,B), atm(A,C,c,27,D).
\end{verbatim}
\end{small}

ALEPH++ExactL1  instead learned from all the folds 
\begin{small}
\begin{verbatim}
active(A) :- atm(A,B,c,27,C), lumo(A,D), lteq(D,-1.749).
\end{verbatim}
\end{small}

The Progol clauses
\begin{small}
\begin{verbatim}
active(A) :- atm(A,B,h,3,0.149).
active(A) :- atm(A,B,h,3,0.144).
\end{verbatim}
\end{small}
mean that a compound with a hydrogen atom $h$ of type 3 with partial charge {\small 0.149}  or 0.144 is active. Very similar charge values (0.145) are found by {\small ALEPH++ExactL1}. \\
SLIPCOVER learned

\begin{small}
\begin{verbatim}
active(A):0.945784 :- atm(A,B,h,3,C),lumo(A,D),D=<-2.242.
active(A):0.01595  :- atm(A,B,h,3,C),logp(A,D),D>=3.26.
active(A):0.00178048 :- benzene(A,B),ring_size_6(A,C),atm(A,D,h,3,E).
\end{verbatim}
\end{small}
SLIPCASE instead learned clauses that relate the drug activity mainly to benzene compounds and 
energy and charge values; for instance one theory is:

\begin{small}
\begin{verbatim}
active(A):0.299495 :-  benzene(A,B),lumo(A,C),lteq(C,-1.102),benzene(A,D),
         logp(A,E),lteq(E,6.79),gteq(E,1.49),gteq(C,-2.14),gteq(E,-0.781).
active(A) :-  lumo(A,B),	lteq(B,-2.142),lumo(A,C),gteq(B,-3.768),lumo(A,D),
              gteq(C,-3.768).
\end{verbatim}
\end{small}

\paragraph{Hepatitis}

The language bias for SLIPCOVER and SLIPCASE allowed \texttt{type/2} only in the head and all the other predicates in the body of clauses,  hence the depth $D$ was not relevant.
For SLIPCOVER, $NBC$ was not relevant as only  \texttt{type/2} can  appear in the clause heads, and the approximate semantics was necessary to limit learning time.
The initial theory for SLIPCASE contained the two facts  \verb|type(A,type_b):0.5.| and \verb|type(A,type_c):0.5.|

For LSM, we used the  discriminative  training  algorithm for learning the weights, by  specifying \texttt{type/2} as the only non-evidence predicate, and
the MC-SAT algorithm for inference over the test fold, by  specifying \texttt{type/2} as the query predicate.
\vspace{0.1cm}

SLIPCOVER achieves significantly higher AUCPR and AUCROC than SLIPCASE and LSM.


Examples of clauses learned by SLIPCOVER are
\begin{small}
\begin{verbatim}
type(A,type_b):0.344348 :- age(A,age_1).
type(A,type_b):0.403183 :- b_rel11(B,A),	fibros(B,C).
type(A,type_c):0.102693 :- b_rel11(B,A),	fibros(B,C),b_rel11(D,A),
                           fibros(D,C),age(A,age_6).
type(A,type_c):0.0933488 :- age(A,age_6).
type(A,type_c):0.770442 :- b_rel11(B,A),	fibros(B,C),b_rel13(D,A).					
\end{verbatim}
\end{small}

Examples of clauses learned by SLIPCASE are
\begin{small}
\begin{verbatim} 
type(A,type_b):0.210837.
type(A,type_c):0.52192 :- b_rel11(B,A),fibros(B,C),b_rel11(D,A),fibros(B,E).
type(A,type_b):0.25556.
\end{verbatim}
\end{small}

LSM long execution time is mainly affected by the \textit{createrules} phase, where LSM counts the true groundings
of all possible unit and binary clauses to find those that are always true in the data: it took 17 hours on all folds; moreover this phase produces only one short clause in every fold. %

\begin{table}[ht]
    \caption{Results of the experiments in terms of the  Area Under the PR Curve averaged over the folds.  The standard deviations are also shown. }
    \label{tab:pr}
    \begin{minipage}{\textwidth}
\begin{tabular}{lccccc}
\hline \hline
System &HIV &UW-CSE &WebKB &Mutagenesis &Hepatitis\\ \hline
SLIPCOVER
&$0.82 \pm0.05$ &$0.13\pm0.02$ &$0.47\pm0.05$ &$0.95\pm0.01$ & $0.80\pm0.01$

\\
SLIPCASE
&$0.78\pm0.05$  &$0.03\pm 0.01$ &$0.31\pm0.21$ &$0.92\pm 0.08$ &$0.71\pm0.05$
\\
LSM
&$0.37\pm0.03$ &$0.07\pm0.02$ &- &-  &$0.53\pm 0.04$
\\
SEM-CP-logic
&$0.58 \pm 0.03$  &-  &-  &- &-
\\
Aleph
&-  &$0.07\pm0.02$ &$0.15\pm0.05$  &$0.73\pm0.09$  &-
\\
ALEPH++
& -  &$0.05\pm0.006$  &$0.37\pm0.16$ &$0.95\pm0.009$ &- 
\\
\hline \hline
\end{tabular}
\end{minipage}
\end{table}

\begin{table}[ht]
    \caption{Results of the experiments in terms of the  Area Under the ROC Curve averaged over the folds. The standard deviations are also shown. }
    \label{tab:roc}
    \begin{minipage}{\textwidth}
\begin{tabular}{lcccccc}
\hline \hline
System &HIV &UW-CSE &WebKB &Mutagenesis &Hepatitis\\ \hline
SLIPCOVER
&$0.95\pm0.01$ &$0.93\pm0.01$	 &$0.76\pm0.01$	 &$0.89\pm0.05$ &$0.74\pm0.01$
\\
SLIPCASE
&$0.93\pm0.01$ &$0.89\pm0.03$  &$0.70\pm0.03$  &$0.87\pm0.05$ &$0.66\pm0.06$
\\
LSM 
&$0.60\pm0.003$	 &$0.85\pm0.21$ & -  & - & $0.52 \pm0.06$
\\
SEM-CP-Logic
&$0.72\pm 0.02$ & -  & - & - & -
\\
Aleph
& - &$0.55\pm0.001$  &$0.59\pm0.04$  &$0.53\pm0.04$ & -
\\
ALEPH++
& - &$0.58\pm0.07$  &$0.73\pm0.27$  &$0.90\pm0.004$ & - 
\\
\hline\hline
\end{tabular}
\end{minipage}
\end{table}

\begin{table}[ht]
    \caption{Normalized Area Under the PR Curve for the high-skew datasets. The skew is the proportion of positive examples on the total testing examples.}
    \label{tab:npr}
    \begin{minipage}{\textwidth}
\begin{tabular}{lcc}
\hline \hline
System &Mutagenesis &Hepatitis\\ \hline
Skew	  &0.66				&0.5\\
SLIPCOVER
&0.91 &0.71
\\
SLIPCASE
&0.86  &0.58
\\
LSM
&- &0.32
\\
Aleph
&0.51 &-
\\
ALEPH++
&0.91  &-
\\
\hline \hline
\end{tabular}
\end{minipage}
\end{table}

\begin{table}
    \caption{Execution time in hours of the experiments on all datasets.} 
    \label{tab:times}
    \begin{minipage}{\textwidth}
\begin{tabular}{lcccccc}
\hline \hline
System &HIV &UW-CSE &WebKB &Mutagenesis &Hepatitis\\ \hline
SLIPCOVER &0.115 & 0.067 & 0.807 & 20.924 & 0.036\\  %
SLIPCASE & 0.010 & 0.018 & 5.689 & 1.426 & 0.073\\
LSM & 0.003 & 2.653 & - & - &25\\
Aleph & - & 0.079 & 0.200 & 0.002 & - \\
ALEPH++ & - & 0.061 & 0.320 & 0.050 & -\\
\hline\hline
\end{tabular}
\end{minipage}
\end{table}

\begin{table}
    \caption{Results of t-test on all datasets relative to AUCPR. p is the p-value of a paired two-tailed t-test between \emsalgorithm{} and the other systems (significant differences in favor of \emsalgorithm{} at the 5\% level  in bold).} %
    \label{tab:p-valuePR}
    \begin{minipage}{\textwidth}
\begin{tabular}{lcccccc}
\hline \hline
System  &HIV &UW-CSE &WebKB &Mutagenesis &Hepatitis\\ \hline
SLIPCASE & \textbf{0.02} & 0.08 & 0.24 & 0.15 & \textbf{0.04}\\  %
LSM & \textbf{4.11e-5} &\textbf{0.043} & - & - & \textbf{3.18e-4}\\
SEM-CP-logic & \textbf{4.82e-5} & - & -& - &  -\\
Aleph & - &\textbf{9.48e-4} & 0.06 & \textbf{2.84e-4} &-\\
ALEPH++ExactL1 & - &0.07 & 0.57 & 0.90 & -\\
\hline\hline
\end{tabular}
\end{minipage}
\end{table}

\begin{table}
    \caption{Results of t-test on all datasets relative to AUCROC. p is the p-value of a paired two-tailed t-test between \emsalgorithm{} and the other systems (significant differences in favor of \emsalgorithm{} at the 5\% level  in bold).} 
    \label{tab:p-valueROC}
    \begin{minipage}{\textwidth}

\begin{tabular}{lcccccc}
\hline \hline
System  &HIV &UW-CSE &WebKB &Mutagenesis &Hepatitis\\ \hline
SLIPCASE & \textbf{0.008} & \textbf{0.003} & 0.14 & 0.49 & \textbf{0.050}\\
LSM & \textbf{2.52e-5} & 0.40 & - & - & \textbf{0.003}\\
SEM-CP-logic & \textbf{6.16e-5} & - & - & - & - \\
Aleph & - & \textbf{1.27e-4} & 0.11 & \textbf{3.93e-5} & -\\
ALEPH++ExactL1 & - & \textbf{6.39e-4} & 0.88 & 0.66 & -\\
\hline\hline
\end{tabular}
\end{minipage}
\end{table}

\paragraph{Overall Remarks} \hspace{0pt}
The results in Tables \ref{tab:pr} and \ref{tab:roc} show that \emsalgorithm{} achieves larger areas than all the other systems in both AUCPR and AUCROC, for all datasets except Mutagenesis, where ALEPH++ExactL1 behaves slightly better.

SLIPCOVER always outperforms SLIPCASE due to the more advanced language bias and search strategy. We experimented with various SLIPCASE parameters in order to obtain an execution time similar to SLIPCOVER's and the best match we could find is the one shown. Increasing the number of SLIPCASE iterations often gave a memory error when building BDDs so we could not find a closer match.

Both SLIPCOVER and SLIPCASE always outperform Aleph, showing that a probabilistic ILP system can better model the domain than a purely logical one. 

SLIPCOVER's advantage over LSM lies in a smaller memory footprint, that allows it to be applied in larger domains, and in the effectiveness of the bottom clauses in guiding the search, in comparison with the more complex clause construction process in LSM.

SLIPCOVER improves on ALEPH++ExactL1 by being able to learn disjunctive clauses and by more tightly combining the structure and parameter searches.

The area differences between \emsalgorithm{} and the other systems are statistically significant in its favor in 17 out of 30 cases at the 5\% significance level.

\section{Conclusions}
\label{conc}
We  presented SLIPCOVER, an algorithm for learning both the structure and the parameters of Logic Programs with Annotated Disjunctions by performing a beam search in the space of clauses and a greedy search in the space of theories. It can be applied to all languages that are based on the distribution semantics.

The code of \emsalgorithm{} is available in the source code repository of the development version of Yap and is
published at  \url{http://sites.unife.it/ml/slipcover} together with an user manual.

We tested the algorithm on the real datasets HIV, UW-CSE, WebKB,  Mutagenesis and Hepatitis and evaluated its performance - in comparison with the systems SLIPCASE, SEM-CP-logic, LSM, Aleph and ALEPH++ExactL1 - through the AUCPR and AUCROC, and AUCNPR on Mutagenesis and Hepatitis. \emsalgorithm{} achieves the largest values under all metrics in most cases. This shows that the application of well-known ILP and PLP techniques to the SRL field gives results that are competitive or superior to the state of the art.

 In the future we plan to experiment with other search strategies, such as local search in the space of refinements. Moreover, we plan to investigate whether the techniques of LHL and LSM can help improving the performance.

\bibliographystyle{acmtrans}
\bibliography{bib}
\end{document}